\documentclass{article} 
\usepackage{iclr2026_conference,times}

\usepackage{amsmath,amsfonts,bm}

\def\eqref#1{equation~\ref{#1}}

\def\1{\bm{1}}

\DeclareMathAlphabet{\mathsfit}{\encodingdefault}{\sfdefault}{m}{sl}
\SetMathAlphabet{\mathsfit}{bold}{\encodingdefault}{\sfdefault}{bx}{n}

\usepackage{hyperref}
\usepackage{url}

\usepackage{booktabs}       
\usepackage{amsfonts}       
\usepackage{microtype}      
\usepackage{amssymb}
\usepackage{pifont}%
\usepackage{multirow}
\usepackage{multicol}
\usepackage{nicefrac}
\usepackage{bm}
\usepackage{graphicx}
\usepackage{colortbl}
\usepackage{amsmath, amssymb, amsthm}
\usepackage{booktabs}
\usepackage{siunitx}
\usepackage{graphicx}
\usepackage{xcolor}
\usepackage{multicol}
\usepackage{colortbl}
\usepackage{tabularx}
\usepackage{makecell}
\usepackage{multirow}
\usepackage{diagbox}
\usepackage{enumitem}
\usepackage{algorithm, algorithmic}
\usepackage{tgcursor}
\usepackage{amsmath}
\usepackage{amssymb}
\usepackage{url}
\usepackage{enumitem}
\usepackage{caption}
\usepackage{tocloft}

\usepackage[table]{xcolor}
\definecolor{mygray}{RGB}{230,230,230}
\usepackage{subcaption}
\usepackage{xcolor} 

\title{DiffSparse: Accelerating Diffusion Transformers with Learned Token Sparsity}

\author{Haowei Zhu\textsuperscript{\rm 1,2}, Ji Liu\textsuperscript{\rm 1}, Ziqiong Liu\textsuperscript{\rm 1}, Dong Li\textsuperscript{\rm 1}, Junhai Yong\textsuperscript{\rm 2}, Bin Wang\textsuperscript{\rm 2,3}\thanks{Corresponding author.}, Emad Barsoum\textsuperscript{\rm 1} \\
\textsuperscript{\rm 1} Advanced Micro Devices, Inc. 
\textsuperscript{\rm 2} Tsinghua University
\textsuperscript{\rm 3} BNRist
\\
\texttt{\{haoweiz, liuji, ziqioliu, d.li, ebarsoum\}@amd.com}, \\
\texttt{yongjh@tsinghua.edu.cn}, \texttt{wangbins@tsinghua.edu.cn}
}

\iclrfinalcopy 
\begin{document}

\maketitle

\begin{abstract}
Diffusion models demonstrate outstanding performance in image generation, but their multi-step inference mechanism requires immense computational cost. Previous works accelerate inference by leveraging layer or token cache techniques to reduce computational cost. However, these methods fail to achieve superior acceleration performance in few-step diffusion transformer models due to inefficient feature caching strategies, manually designed sparsity allocation, and the practice of retaining complete forward computations in several steps in these token cache methods.
To tackle these challenges, we propose a differentiable layer-wise sparsity optimization framework for diffusion transformer models, leveraging token caching to reduce token computation costs and enhance acceleration. Our method optimizes layer-wise sparsity allocation in an end-to-end manner through a learnable network combined with a dynamic programming solver. Additionally, our proposed two-stage training strategy eliminates the need for full-step processing in existing methods, further improving efficiency.
We conducted extensive experiments on a range of diffusion-transformer models, including DiT-XL/2, PixArt-$\alpha$, FLUX, and Wan2.1. Across these architectures, our method consistently improves efficiency without degrading sample quality. For example, on PixArt-$\alpha$ with 20 sampling steps, we reduce computational cost by $54\%$ while achieving generation metrics that surpass those of the original model, substantially outperforming prior approaches. These results demonstrate that our method delivers large efficiency gains while often improving generation quality.

\end{abstract}    
\section{Introduction}

In recent years, diffusion models have made remarkable progress in the field of image generation. Among them, the Stable Diffusion series~\citep{rombach2022high,podell2023sdxl, tian2024u, esser2024scaling} has achieved significant success in high-quality image generation~\citep{zhu2024distribution, zhu2025recon}. This advancement is largely attributed to the effectiveness of diffusion probabilistic models (DPM)~\citep{ho2020denoising} and the powerful U-Net~\citep{ronneberger2015u} architecture, which allows high resolution synthesis with detail preservation.
Additionally, some recent works~\citep{peebles2023scalable,chen2023pixartalpha, tian2024u} have explored the integration of diffusion models with Transformer-based architectures, demonstrating outstanding performance. In particular, scaling laws have been leveraged to expand the model size of Transformers~\citep{vaswani2017attention}, further enhancing precision and generative quality, these large-scale models leverage improved expressivity and enhanced generalization, pushing the boundaries of generative artificial intelligence.

However, despite these advances, the substantial computational cost associated with diffusion models presents a significant challenge for real-world deployment. The inference of such large models requires extensive computational resources, which can hinder practical applications. Addressing this issue requires innovations in model acceleration techniques to enable broader accessibility and usability of diffusion-based generative models. 
Existing methods of diffusion model acceleration typically focus on sampler optimization ~\citep{song2020denoising, lu2022dpm}, model pruning ~\citep{structural_pruning_diffusion, zhang2024laptop, fang2023structural}, distillation ~\citep{yin2024stepdistillation, luo2023latent, salimans2022progressive}, and feature caching \citep{selvaraju2024fora, ma2024deepcache, liu2025timestep, liu2025reusing}. 
Feature caching methods leverages temporal redundancy to reuse intermediate features, achieving significant speedups. They become popular in the field of diffusion model acceleration due to non-training diffusion model and easy integrating into the original inference pipeline. 
Previous methods cache and reuse coarse-grained, layer-level features, whereas token cache methods~\citep{zou2025accelerating, duca, zhang2025training} reuse token-level features, achieving better acceleration performance. However, these approaches require manual sparsity allocations and hand-crafted schedules that preserve several full forward passes during denoising, which limits the acceleration potential of token-level feature caching.

To address these challenges, we propose DiffSparse, a learnable framework for optimizing layer-wise sparsity allocation in diffusion transformer models. Our approach dynamically determines the optimal sparsity configuration across all layers and inference steps, ensuring that the overall pruning rate is met in an end-to-end manner through a model-driven process. Moreover, DiffSparse eliminates the need for complete forward computations in predefined steps required by existing methods, further enhancing efficiency.

Specially,  our approach formulates the token cache optimization as a dynamic programming-based sparsity allocation problem.  We innovatively design a learnable sparsity cost predictor, which predicts a cost matrix that quantifies the sparsity costs associated with target sparsity rates for all layers across every denoising step. Then we propose a dynamic programming approach to determine the optimal sparsity configurations for all layers over the relevant denoising steps, minimizing the overall sparsity cost while satisfying the required sparsity rate. Finally, we introduce a token selector that dynamically selects a specific proportion of tokens for reuse, leveraging the learned sparsity ratio to accelerate inference. To optimize the learnable sparsity cost predictor, we utilize a perceptual distillation loss that minimizes the degradation in generation quality.
Furthermore, we introduce a two-stage training strategy that eliminates the need for complete forward computations in predefined steps required by existing methods while also improving accuracy. We have conducted extensive experiments on various transformer-based baselines, and the pruning results outperform other SOTA pruning methods by a large margin. 
For example, pruning $54\%$ of tokens yields an FID of 27.79, our method substantially better than the state-of-the-art methods ToCa (28.35) and TaylorSeer (29.08), while achieving a higher speedup (1.91$\times$) on PixArt-$\alpha$. These results underscore the practical effectiveness of our method. Our contributions are summarized as follows:
\begin{itemize}
\item We propose DiffSparse, a differentiable approach to optimize layer-wise token sparsity in diffusion models sampling process. By integrating a sparsity cost predictor, dynamic programming solver, and adaptive token selector, it automates sparsity allocation and token reuse without manual heuristics.
\item We introduce a two-stage training strategy that eliminates the need for predefined complete forward computations in several steps required by existing methods, fully unlocked the acceleration potential of token-level feature caching.
\item Extensive experiments on diverse foundation models prove that our method surpasses existing SOTA methods by a large margin, setting new efficiency-accuracy benchmarks.
\end{itemize}
\section{Related Work}

\noindent \textbf{Diffusion Transformer Models.}
The integration of transformers into diffusion models has significantly advanced generative modeling, improving scalability and performance. Diffusion models, which generate data by iteratively denoising from a noise distribution. Traditionally, diffusion models relied on CNNs, but recent studies demonstrate the effectiveness of transformers \citep{peebles2023scalable, chen2023pixartalpha, tian2024u, openai2024sora, chen2024pixart, wu2025drivescape}. Diffusion Transformer (DiT)~\citep{peebles2023scalable} replaces the U-Net backbone with a transformer, leveraging long-range dependencies and efficient scaling to achieve superior image generation. PixArt~\citep{chen2023pixartalpha} builds on this by introducing a hierarchical transformer architecture and a novel noise schedule, excelling in high-resolution and text-to-image synthesis.
Although diffusion transformer models have achieved great success, the substantial computational overhead from the iterative denoising process makes them inefficient for industrial deployment.

\paragraph{Acceleration of Diffusion Models.}
 Diffusion acceleration is a critical research area focused on reducing computational costs and improving inference efficiency while preserving high-quality generation. Recent advancements can be categorized into sampler optimization \citep{songDDIM, lu2022dpm, lu2022dpm++}, model pruning \citep{structural_pruning_diffusion, zhang2024laptop}, distillation \citep{ salimans2022progressive, yin2024improved}, and feature caching \citep{li2023FasterDiffusion, ma2024deepcache, zhu2025dip}. 
  Sampler optimization reduces the number of denoising steps during inference using deterministic or adaptive strategies to approximate the denoising process efficiently. 
 Model pruning removes redundant parameters and achieving speedups with structured pruning~\citep{fang2023structural}. 
 Other strategies, such as Rectified Flow \citep{liu2022flow} and knowledge distillation~\citep{yin2024improved} accelerates inference by matching model outputs in fewer steps without quality loss.

Feature caching is particularly effective for DiT architectures. 
Methods such as FORA \citep{selvaraju2024fora} and $\Delta$-DiT \cite{chen2024delta-dit} reuse attention and MLP representations, while DiTFastAttn \citep{yuan2024ditfastattn} further reduces redundancies in self-attention. 
Dynamic strategies like TeaCache \citep{liu2025timestep} estimate timestep-dependent differences, and 
TaylorSeer \citep{liu2025reusing} introduced a ``cache-then-forecast'' paradigm that predicts and updates cached features, though its advantage is most evident with long-range caching. SpeCa \citep{liu2025speca} and TAP  \citep{zhu2026tap} further enhance the performance with speculative sampling and adaptive prediction. 
Complementary to these are token cache methods~\citep{zou2025accelerating, duca, zhang2025training, you2025layer}, which apply fine-grained, error-guided token-wise caching to dynamically update features, achieving substantial acceleration without compromising quality.
{More discussion with existing methods are presented in Appendix.}

In this paper, we introduce DiffSparse, a feature‐caching approach for accelerating diffusion transformer models. These models typically require only a few dozen sampling steps and have seen growing adoption in industry. 
Unlike prior works~\citep{zou2025accelerating, duca}, DiffSparse employs a token‐level cache within an end‐to‐end learning framework that casts model acceleration under a fixed compression ratio as a layer‐wise sparsity optimization problem across timesteps, eliminating the need for manually tuned sparsity or acceleration parameters. To address inefficiencies in existing approaches, which depend on predefined full‐step computation schedules, we also propose a two‐stage training protocol that adaptively allocates computation where it is most needed.

 \section{Method}

 In this section, we start with a brief introduction to the diffusion transformer model and the token cache strategy. We then present the challenges of the existing token caching approaches. Finally, we present our DiffSparse approach, which builds upon the token cache strategy for acceleration and optimizes the layer-wise token sparsity of diffusion transformer model in a learnable manner, enhancing accuracy while maintaining the sparsity requirement.

\subsection{Preliminary}

\noindent \textbf{Diffusion Models.} 
Diffusion models are a class of generative models that construct a Markov chain of latent variables by progressively adding Gaussian noise to data samples and then reversing this process to synthesize new samples. Given an initial data sample \( x_0 \), the forward diffusion process transforms the data through a series of steps:
\begin{equation}
q(x_t \mid x_{t-1}) = \mathcal{N}\left(x_t; \sqrt{1-\beta_t}\, x_{t-1},\, \beta_t \mathbf{I}\right), 
\end{equation}
where $t$ is the time step, \(\{\beta_t\}_{t=1}^T\) denotes a predefined variance schedule. After \(T\) steps, the data is nearly transformed into an isotropic Gaussian distribution, i.e., \(q(x_T) \approx \mathcal{N}(0, \mathbf{I})\).

The reverse process is parameterized by a noise prediction network, which aims to recover the original data by iteratively removing the added noise, and is modeled as:
\begin{equation}
p_\theta(x_{t-1} \mid x_t) = \mathcal{N}\left(x_{t-1}; \mu_\theta(x_t, t),\, \Sigma_\theta(x_t, t)\right),
\end{equation}
where \(\mu_\theta\) and \(\Sigma_\theta\) are learned functions. Because the network is applied at each timestep in the multi-step denoising process, the repeated evaluations of the noise prediction network dominate the computational cost, accounting for the majority of the model's floating-point operations (FLOPs).

\paragraph{Diffusion Transformer.}
The Diffusion Transformer~\citep{chen2023pixartalpha}  is a novel architecture that synergizes the iterative refinement capabilities of diffusion processes with the representational power of transformers. In this framework, the input is represented as a set of tokens \(\mathbf{X} \in \mathbb{R}^{N \times D}\), where \(N\) denotes the number of tokens and \(D\) their dimensionality. The network architecture is composed of \(L\) stacked blocks, each integrating three key components: a self-attention (SA) layer, a cross-attention (CA) layer, and a multi-layer perceptron (MLP) layer.
The self-attention mechanism enables the model to capture long-range dependencies among tokens. In parallel, the cross-attention module facilitates the incorporation of conditioning information, enhancing the model's ability to generate contextually relevant outputs. The subsequent MLP further refines these token representations through non-linear transformations.

A significant advantage of the Diffusion Transformer lies in its ability to iteratively refine token representations during the denoising process, leading to improved sample quality. This layered approach allows the model to effectively balance global context and local details, thereby offering enhanced performance in complex generative tasks.

\paragraph{Token-Wise Feature Caching Approach.}
Prior work~\citep{ma2024deepcache} has demonstrated that features at adjacent timesteps exhibit high similarity, leading to significant redundancy. To exploit this redundancy for computational efficiency, previous approaches~\citep{ma2024deepcache,wimbauer2024cache} have introduced caching mechanisms that reuse features to accelerate processing. The token-wise feature caching approach~\citep{zou2025accelerating} operates at a finer granularity by caching features at the individual token level, enabling more effective exploitation of the redundancy.

Token-wise feature caching mechanism begins by computing and storing the intermediate token features  \(\mathbf{X} = \{\hat{x}_0, \hat{x}_1, \ldots, \hat{x}_{N-1}\}\) from each self-attention, cross-attention, and MLP layer into a cache \( C \) at the initial timestep \( t \).
In subsequent timesteps, a predefined cache ratio 
\( R \) determines the proportion of tokens reused from the cache 
\( C \)  for each layer at each timestep. The \( R \) selected tokens based on token importance rank, denoted as \( I_{\text{Cache}} \), will bypass re-computation by reusing their cached values, while the remaining tokens \( I_{\text{Compute}} = \{\hat{x}_i\}_{i=1}^{N} \setminus I_{\text{Cache}} \) are recomputed. For a given layer \( f \), the computation for each token \( \hat{x}_i \) is formulated as:
\begin{equation}
F(\hat{x}_i) = \gamma_i f(\hat{x}_i) + (1-\gamma_i) C(\hat{x}_i),
\end{equation}
where \(\gamma_i = 0\) for \( \hat{x}_i \in I_{\text{Cache}} \) and \(\gamma_i = 1\) for \( \hat{x}_i \in I_{\text{Compute}} \). To mitigate error accumulation from reused features, the cache is dynamically updated for tokens in \( I_{\text{Compute}} \) via:
\begin{equation}
C(\hat{x}_i) \leftarrow F(\hat{x}_i).
\end{equation}

This token-wise feature caching approach effectively reduces redundant computations by leveraging the high similarity of features across adjacent timesteps, thus significantly accelerating the inference process while maintaining robust feature representations.

\paragraph{Challenges in Existing Token Caching Approaches.}
While token caching methods~\citep{zou2025accelerating} have shown great promise in speeding up diffusion transformers, key limitations remain. First, they require manually setting a reuse sparsity rate for each layer at every timestep, resulting in a large, hard-to-tune parameter space. This manual process hampers performance and scalability. A learnable or adaptive sparsity strategy could unlock further gains.
Second, current methods still depend on a full-step design (several steps without caching) to maintain generation quality. However, this compromises the efficiency of token-based operations. Replacing this with dynamic caching tailored to diffusion transformers can better balance quality and speed.
In this paper, we propose an intelligent framework that jointly learns optimal sparsity across layers and removes the reliance on full-step computation, significantly improving both performance and flexibility.

\begin{figure*}[ht]
    \centering
    \includegraphics[width=0.8\linewidth]{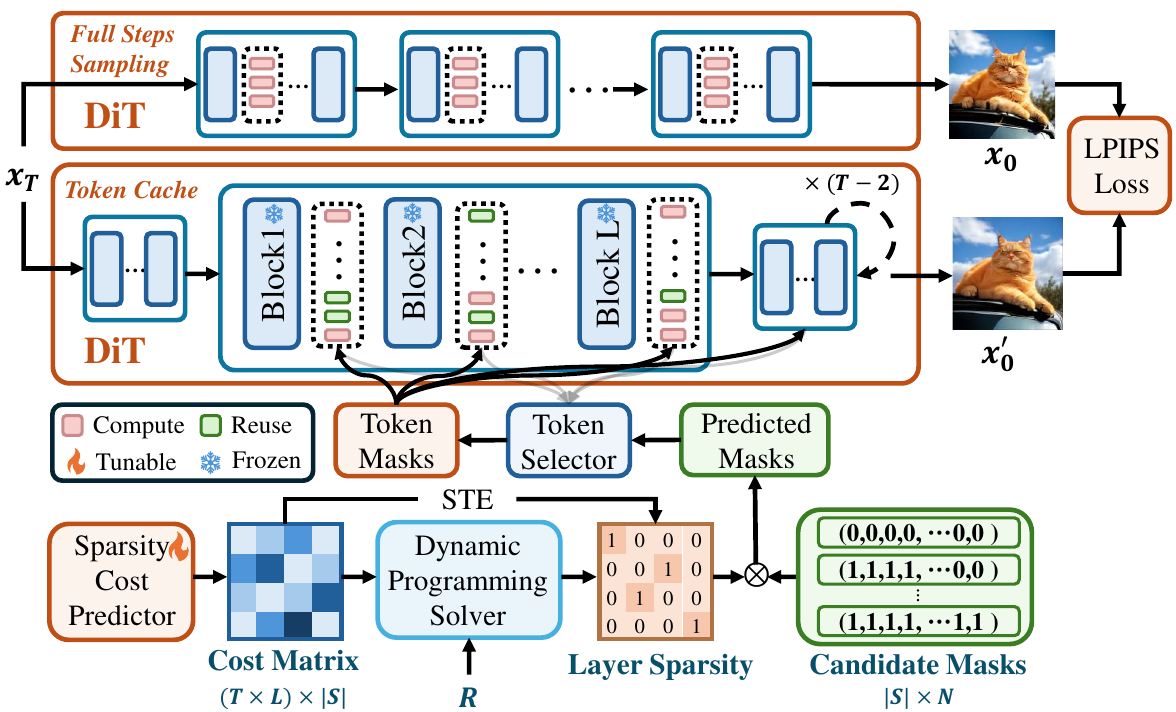}
    \caption{DiffSparse uses a learnable sparsity-cost predictor and dynamic programming to learn per-layer sparsity under target ratio \(R\). We generate binary masks from the chosen sparsity maps and candidate masks. A token selector reuses features from previous diffusion steps to skip unimportant tokens and speed sampling. To enable gradient flow through the binary masks, we apply Straight-Through Estimation (STE) and train our model using full-step sampling targets with LPIPS loss.}

    \label{fig:overview}
    \vspace{-4mm}
\end{figure*}

\subsection{DiffSparse Approach}

To automate per-layer sparsity selection and remove the reliance on full-step designs, we propose DiffSparse, an efficient token caching framework for diffusion transformers. DiffSparse learns layer-wise sparsity end-to-end by combining a learnable sparsity cost predictor with a dynamic programming solver to find optimal sparsity configurations across layers and denoising steps. It also adopts a two-stage training scheme that gradually replaces full computation steps with cache-based ones, improving efficiency without sacrificing performance.
As illustrated in Figure~\ref{fig:overview}, DiffSparse comprises three components: a token selector, a sparsity cost predictor, and a dynamic programming solver. The cost predictor estimates a cost matrix representing the sparsity cost for various predefined rates across all layers and denoising steps (excluding the first). The dynamic solver then identifies the optimal sparsity pattern under a global sparsity constraint \( R \). Based on this, the token selector determines which tokens to reuse and which to recompute at each layer. Training is guided by a perceptual distillation loss, integrated into a two-stage training pipeline for effective learning.

\paragraph{Token Selector.}
We employ a \emph{Token Selector} that assigns each token $\hat{x}_i$ an importance score used to decide which tokens are freshly computed and which remain cached. The score is a composite, layer-wise quantity of the form:

\begin{equation}
  S(\hat{x}_i) \;=\; B\!\Big(\sum_{{q}=1}^{{Q}} \lambda_{{q}}\, s_{{q}}(\hat{x}_i)\Big),
  \label{eq:token_selector}
\end{equation}

where each $s_{{q}}(\hat{x}_i)$ is a scalar signal capturing a different criterion (for example, self-attention influence, cross-attention focus, cache-reuse frequency, \emph{etc.}), and 
$\{\lambda_{{q}}\}_{{q}=1}^{{Q}}$ are weighting hyperparameters that balance these criteria. The operator $\mathcal{B}(\cdot)$ is optional and denotes a spatial bonus operation that promotes a spatially uniform coverage of selected tokens (implemented, e.g., by boosting tokens that are local maxima within a $k\times k$ neighborhood). Other choices for \(\mathcal{B}\) are possible (e.g. smooth kernels or distance-based adjustments).

Given the per-token scores $S(\hat{x}_i)$ in a layer with $N$ tokens, we sort tokens by descending score and select the top $K$ tokens according to a predefined sparsity ratio $R$. 
We emphasize that our contribution is orthogonal to any particular token-ranking heuristic: the choice of scoring components (e.g. self-attention influence, cross-attention terms, spatial bonus) is optional and can be replaced by alternative ranking methods. Detailed descriptions and comparisons of specific token-ranking strategies are provided in the Appendix \ref{appendix:token_selector}. Empirically, our allocation scheme yields consistent gains across different token-ranking methods (see Table~\ref{table: Ablation-TokenSelection}).

\paragraph{Learnable Sparsity Cost Predictor.}
We propose a learnable sparsity cost predictor to adaptively determine layer-wise sparsity in diffusion transformers (DiTs) while balancing inference efficiency and computational cost. Given a DiT with \(L\) layers operating over \(T\) denoising timesteps, our goal is to generate a binary mask \(M \in \{0,1\}^N\) for each layer \(l\) and timestep \(t\) that selects \(K_{l,t}\) tokens for full computation and reuses features for the remaining \(N-K_{l,t}\) tokens. This is formalized as a constrained optimization over a candidate sparsity set \(S\), where \(|S|\) denotes the number of predefined sparsity configurations.
For a layer containing \(N\) tokens, let \(S\) denote the set of sparsity rates, each a value between 0 and 1, at which we retain a corresponding fraction of tokens. For instance, if \(N=256\) and we choose a step size of 32 tokens, we obtain $S = \{0,0.25,0.50,0.75,1.0\}$, which corresponds to retaining \(\{0,\,64,\,128,\,192,\,256\}\) tokens, respectively. 
Our objective is to learn the relative cost of applying different sparsity rates across layers. Our experimental results (Table~\ref{tab:pixart-512}) demonstrate that the learned sparsity predictor generalizes across resolutions, so that a sparsity allocation trained at low resolution remains effective at higher resolutions.

We implement the sparsity cost predictor using \((T \times L) \times |S|\) learnable parameters, where \(T\) is the number of timesteps, \(L\) is the number of layers, and \(|S|\) is the size of the candidate sparsity set.
The predictor outputs a normalized cost matrix \( C \in \mathbb{R}^{(T \times L) \times |S|} \), where each entry \( C_{(t,l),s} \) quantifies the cost of applying sparsity configuration \(s \in S\) to layer \(l\) at timestep \(t\). We minimize the cumulative cost while ensuring the total sparsity meets a predefined overall pruning rate \( R \). The sorted token set \(\bar{\mathbf{X}} \in \mathbb{R}^{N \times D}\) enables efficient mask selection by prioritizing tokens with high scores.

Importantly, The cost predictor's size depends only on \(T\), \(L\), and \(|S|\), not on token-sequence length $N$. Empirically, we found that simply increasing \(|S|\) beyond a moderate size yields diminishing or negative returns (Table~\ref{table: Sparse-Interval}), and experiments show the learned cost predictor transfers across resolutions (Table~\ref{tab:pixart-512}), demonstrating scalability to high resolutions and robustness to token-length variation.

\paragraph{Dynamic Programming Solver.}
To determine the optimal sparsity configuration while satisfying a global sparsity constraint, we employ a dynamic programming approach to minimize the overall cost across layers. Formally, we define the state function:
\begin{equation}
F(\hat{l}, r) = \min_{\{s_i\}_{i=1}^{\hat{l}}} \sum_{i=1}^{\hat{l}} C_{i, s_i}, \quad \text{s.t.} \quad \sum_{i=1}^{\hat{l}} s_i = r,
\end{equation}
where \( F(\hat{l}, r) \) represents the minimum achievable cost when assigning sparsity levels to the first \( \hat{l} \) layers under a total sparsity constraint \( r \). The recursive formulation is given by:
\begin{equation}
F(\hat{l}, r) = \min_{s \in S, s \leq r} \big( F(\hat{l}-1, r - s) + C_{\hat{l}, s} \big).
\end{equation}
Here, the transition considers all possible sparsity levels \( s \) that can be allocated to layer \( \hat{l} \), ensuring that the total sparsity constraint is maintained. The algorithm iteratively computes \( F(\hat{l}, r) \) for \( \hat{l} = 1, \dots, L \cdot T \) and \( r = 0, \dots, \hat{R} \), followed by a backtracking step to reconstruct the optimal sparsity allocation, where \( \hat{R} = R \cdot L \cdot T \). This approach operates with a time complexity of \( O((L \cdot T)^2 \cdot |S|) \), making it computationally feasible for practical deep learning scenarios. To reduce the number of redundant state computations and lower overall complexity, we implement pre-pruning strategies. 
For example, when target sparsity ratio \(R = 43\%\), \(|S| = 5\), \(T = 20\), and \(L = 28\), it requires about 4 hours {(including DP optimization and fine-tuning)} of total training time. 
{
The DP solver runs in approximately $\approx$30 seconds for the configurations reported, but it is not executed at inference time. At inference, the model only uses the precomputed masks.
}
Since the direct conversion of the predicted cost matrix \(C\) to a discrete mask \(M\) is non-differentiable, we utilize the Straight-Through Estimator (STE) \citep{jang2016categorical} to approximate the gradients of the discrete mask with respect to the cost predictions. This approach facilitates end-to-end optimization of the sparsity cost predictor. 

\paragraph{Training Loss.}
To guide the optimization of the pruned Diffusion Transformer, we employ the Learned Perceptual Image Patch Similarity (LPIPS) loss \citep{zhang2018lpips} as a perceptual distillation loss. In our framework, the original model prior to token pruning serves as the teacher network, while the pruned model is treated as the student network. Both models generate outputs via a multi-step sampling process inherent to diffusion models. 

Let  $x_0$ and \( x'_0 \) denote the multi-step sampling outputs from the teacher and student networks, respectively. The LPIPS loss is then defined as:
\begin{equation}
\mathcal{L}_{\text{LPIPS}} = \text{LPIPS}(x_0, x'_0),
\end{equation}
which measures the perceptual similarity between the outputs. During training, gradients are backpropagated solely through the student network, as the teacher network's parameters are detached (i.e., its gradients are not computed). This setup ensures that the student model is effectively distilled to mimic the perceptual characteristics of the teacher model, thereby achieving acceleration through token pruning while preserving output quality.

\paragraph{Two-Stage Training Strategy.}  
We propose a two-stage training framework to optimize the cost matrices for \texttt{full-step positions} and \texttt{layer sparsity} components.
In the first stage, we follow \citep{selvaraju2024fora, zou2025accelerating} to preset \(T_f\) full-step positions and independently optimize the step cost matrix \(C_f \in \mathbb{R}^{T \times 2}\) encoding temporal sparsity decisions  
and the layer sparsity cost matrix \(C_l \in \mathbb{R}^{(L \times T) \times |S|}\) governing token retention per layer. We first solve \(C_f\) via dynamic programming to identify \(|T_f|\) optimal full-step positions with minimal cumulative cost. For these selected steps, we warm-start layer sparsity optimization by subtracting $\delta$ from the predicted costs:  
\begin{equation}
\small
    C_l^{(t,l,s)} \gets C_l^{(t,l,s)} - \delta\ \quad \forall t \in T_f, l \in \{1,...,L\}, s = N.
\label{eq:cost-merge}
\end{equation}

This strategy preserves inter-layer cost ranking while leveraging full-step error correction capabilities.  

In the second stage, we integrate step and layer costs by modifying layer sparsity entries using Equation \ref{eq:cost-merge}. The unified cost matrix is then fine-tuned to systematically redistribute FLOPs across sampling steps. Unlike existing methods \citep{selvaraju2024fora, zou2025accelerating} that rigidly enforce full steps for noise correction, our approach dynamically optimizes sparsity patterns through differentiable cost interaction. The pseudocode is provided in the supplementary materials.


\section{Experiments}
\label{sec:experments_v2}
\subsection{Experiment Settings}
\label{sec:experments_setting_v2}
\paragraph{Model Configurations.}
We conduct experiments on four widely used DiT-based models across various generation tasks: (1) \textbf{PixArt-\(\alpha\)} with 20 DPM Solver++ \citep{lu2022dpm++} steps and \textbf{FLUX.1-schnell} \citep{flux2024} with 4 steps for text-to-image generation and (2) \textbf{DiT-XL/2} with 50 DDIM \citep{songDDIM} steps for class-conditional image generation. (3) \textbf{Wan2.1-1.3B} ~\citep{wan2025wan} with 25 flow-matching sampling steps for text-to-video generation.   
We define the candidate set $S$ as the range from 0 to 1 with an interval of 0.25, yielding $|S| = 5$ token sparsity  candidates.  
More details of the implementation are provided in the supplementary material. 

\paragraph{Training.} 

For PixArt‑$\alpha$ \citep{chen2023pixartalpha} and FLUX.1‑schnell \citep{flux2024}, we train the learnable sparsity‑cost predictor on 10,000  captions randomly sampled from the COCO \citep{lin2014coco} train dataset. For DiT‑XL/2 \citep{peebles2023dit}, we use 10,000 ImageNet \citep{deng2009imagenet} train category indices, and for Wan2.1 we sample 10,000 captions from WebVid-10M \citep{Bain21} for training. During training we use no image data, only captions or class-conditioning information, which do not overlap with the evaluation set. 

We leverage the layer sparsity configuration in the token-cache-based model \citep{zou2025accelerating} to initialize our sparse router training. All the models are trained with AdamW optimizer.
The sparsity cost predictor is trained in two stages. For the first stage, the layer sparsity cost component is optimized for 1 epoch with a learning rate of \(\eta = 1.0\), while the step cost component is trained separately using \(\eta = 0.01\) to capture temporal patterns across denoising steps. For the second stage, we integrate the step cost into the layer-wise costs with \(\delta = 10\) and then fine-tuned for 1 epoch with \(\eta = 0.1\) to optimize layer sparsity allocation. Training requires approximately 4-10 hours on 8 AMD MI250 GPUs with 80GB memory per experiment.

\paragraph{Evaluation.} 
For text-to-image generation, we evaluate on the \textbf{COCO} dataset \citep{lin2014coco} using 30,000 samples at 256 × 256 resolution and \textbf{PartiPrompts} \citep{yu2022scaling} with 1,632 samples. Image quality is quantified by \textbf{FID-30k} \citep{heusel2017fid}, which compares generated images against originals, while text-image alignment is measured by two complementary metrics: \textbf{CLIP-Score} (computed with CLIP-ViT-Large-14 \citep{hessel2021clipscore}) and \textbf{Image Reward} \citep{xu2023imagereward}, a metric shown to more accurately reflect human preferences.
For class-conditional image generation, 50,000 images at 256 $\times$ 256 resolution are generated from 1,000 \textbf{ImageNet} \citep{deng2009imagenet} classes and evaluated using the \textbf{FID-50k} metric.
We evaluate text-to-video generation using the \textbf{VBench} framework on 950 prompts, generating 4,750 videos at 256 $\times$ 256 resolution, each lasting 2 seconds at 8 frames per second, and assess them across \textbf{16 metrics}.

\subsection{Main Results}
\paragraph{Results on Text-to-Image Generation.}
We compare DiffSparse with existing methods under identical sparsity budgets. FORA, {DeepCache (CVPR'24)} and TaylorSeer (ICCV'25) are evaluated with cache interval $N=2$, while {DiCache}, ToCa (ICLR'25) and DuCa are tested using their respective optimal configurations.
Table \ref{table:main_ldm_PixArt} shows that DiffSparse delivers both faster inference and improved generation quality compared with existing methods. At roughly $1.74\times$ speed-up, existing methods suffer degraded image quality, while DiffSparse achieves a strong FID of 26.91 (vs.\ TaylorSeer’s 29.08 and ToCa’s 28.35). {
This corresponds to a relative +5.1\% improvement in FID of DiffSparse over ToCa.}
Pushing further, DiffSparse attains 1.91$\times$ acceleration while producing an FID that surpasses the original (full) model. This improvement stems from a learned sparsity schedule that accelerates convergence of the generated image distribution and improves visual fidelity, while preserving semantic alignment with the conditioning signal.
We provide additional text-to-image comparisons in Appendix~\ref{appendix:more-exp}, and also present more qualitative visual comparison in Appendix~\ref{appendix:qualitative}.

\begin{table*}[htbp]
    \small
    \centering
      \caption{Results of text-to-image generation on MS-COCO2017 with $\text{PixArt-$\alpha$}$ and 20 DPM++ steps.
    }
    \label{table:main_ldm_PixArt}
    \vspace{-8pt}
      \begin{tabular}{l | c c | c c }
        \toprule
        Method  & MACs (T)$\downarrow$ &  Speedup$\uparrow$ & FID-30k$\downarrow$& CLIP$\uparrow$ \\ 
        \midrule

      {PixArt-$\alpha$} \citep{chen2023pixartalpha}   &  2.86  & {1.00$\times$} & 28.20  & 0.163  \\
      \midrule
      {{50\% steps} }  &  1.43  &  {1.74$\times$} & 37.57  & 0.158  \\
      {{FORA $(\mathcal{N}=2)$} \citep{selvaraju2024fora}}     & 1.43 & {1.64$\times$} & 29.67   & 0.164 \\
      {DeepCache (\(\mathcal{N}=2\)) \citep{ma2024deepcache}} & {1.48} & {1.61$\times$} & {29.61} & {0.163} \\
      {DiCache \citep{bu2025dicache}} & {1.63} & {1.77$\times$} & {28.19} & {0.164}  \\
      {ToCa \citep{zou2025accelerating} } &  1.64 & {1.75$\times$} & {28.35} & 0.164   \\
      DuCa \citep{duca}   & 1.63 & {1.78$\times$} & 27.98 & 0.164  \\
      TaylorSeer \citep{liu2025reusing} & 1.57 & 1.83$\times$ & 29.08 & 0.163  \\
    \rowcolor{mygray} {\textbf{{DiffSparse}} (R = 43\%) } &  {1.64} & {1.74$\times$} & \textbf{26.91} & \textbf{0.164} \\
    \rowcolor{mygray} {\textbf{DiffSparse}} (R = 54\%) &  \bf{1.30} &  {\bf{1.91$\times$}} & {27.79} & 0.164  \\
        \bottomrule
      \end{tabular}
  \end{table*}

\begin{table}[ht]
    \centering
    \small
  \caption{Results of class-conditional generation with DiT-XL/2 and 50 DDIM steps on ImageNet.}
    \label{tab:dit}
    \vspace{-8pt}
    \begin{tabular}{l|cc|cccc}
        \toprule
        Method & MACs (T) $\downarrow$  &  Speedup$\uparrow$ & FID$\downarrow$ & sFID $\downarrow$ & Precision $\uparrow$ & Recall  $\uparrow$ \\
        \midrule
        {{$\text{DDIM-50 steps}$}}   & 11.44  & {1.00$\times$} & 2.26 & 4.29 & 0.80 & 0.60 \\
        {{$\text{DDIM-40 steps}$}}   & 9.14   & 1.24$\times$ & 2.39  & 4.28 & 0.80 & 0.59 \\
        {{$\text{DDIM-25 steps}$}}   & 5.73   & 1.96$\times$ & 3.01 & 4.60 & 0.79 & 0.58\\
        {{$\text{DDIM-20 steps}$}}   & 4.58   & 2.42$\times$ & 3.48 & 4.64 & 0.79 & 0.56 \\
        \midrule     
        {{FORA}}  & 4.13   & 2.12$\times$ & 3.88 & 6.74 & 0.79 & 0.56 \\
        {ToCa}      & 4.97  & 2.09$\times$ & 3.05 & 4.70 & 0.79 & 0.57  \\
        {DuCa} & 4.94 & 2.10$\times$ & 3.04 & 4.70 & 0.79 & 0.57 \\
        \rowcolor{gray!20}
        \textbf{DiffSparse } & 4.97 & 2.07$\times$ & \bf{2.81} & \textbf{4.61} & \textbf{0.80} & \textbf{0.59} \\
        \bottomrule
    \end{tabular}
\end{table}

\paragraph{Results on Class-Conditional Image Generation.}
Table~\ref{tab:dit} compares faster sampler DDIM with fewer steps, FORA, ToCa, DuCa and DiffSparse. Our method achieves a better speed–accuracy balance by reallocating computation to the most important layers. At the same acceleration ratio, DiffSparse improves the FID from 3.05 to 2.81, outperforming ToCa by 8\% at 2.07× acceleration. demonstrating its ability to preserve detail and improve image fidelity in diffusion model acceleration. 

\begin{center}
\setlength{\tabcolsep}{6pt}
\small
\vspace{-10pt}
\begin{minipage}[t]{0.54\textwidth}
  \captionof{table}{Comparison in text-to-video generation for Wan2.1-1.3B with 20 sampling steps on VBench.}
  \label{tab:t2v}
  \vspace{-8pt}
  \centering
  \resizebox{\textwidth}{!}{
  \begin{tabular}{l|ccc}
    \toprule
    Method                & MACs (T) $\downarrow$ & Speedup $\uparrow$ & VBench  $\uparrow$\\
    \midrule
    Wan 2.1 - 1.3B              & 43.866   & 1.00$\times$             & 43.82  \\
    {50\% steps} & {21.933} & {1.86$\times$} & {43.14} \\
    DuCa (R = 54\%)   & 20.332   & 1.69$\times$             & 43.56  \\
    DuCa (R = 59\%)   & 18.124   & 1.68$\times$             & 43.30  \\
    \rowcolor{gray!20}
        \textbf{DiffSparse } & 18.124   & \textbf{2.05$\times$}    & \textbf{43.83}  \\
    \bottomrule
  \end{tabular}%
  }
  \vspace{4pt}

  \vspace{-8pt}
\end{minipage}\hfill
\begin{minipage}[t]{0.42\textwidth}
    \captionof{table}{Comparison on PixArt-$\alpha$ using 20 sampling steps at 512$\times$512 resolution.}
  \label{tab:pixart-512}
  \centering
\resizebox{\textwidth}{!}{
  \begin{tabular}{l|ccc}
    \toprule
    Method              & MACs (T) $\downarrow$ & FID $\downarrow$  & CLIP $\uparrow$ \\
    \midrule
    PixArt-$\alpha$     & 10.851   & 21.95            & 0.164           \\ 
    50\% steps            & 5.426    & 25.05            & 0.163           \\
    ToCa                & 5.993    & 23.02            & 0.165           \\
    \rowcolor{gray!20}
        \textbf{DiffSparse }          & 5.986    & \textbf{22.42}   & \textbf{0.165}  \\
    \bottomrule
  \end{tabular}%
}
  \vspace{4pt}

  \vspace{-10pt}
\end{minipage}
\end{center}

\paragraph{Results on Text-to-Video Generation.}
Table~\ref{tab:t2v} presents a comparison between DiffSparse and DuCa~\citep{duca} on Wan2.1-1.3B \citep{wan2025wan} using 20 sampling steps. The methods are comprehensively evaluated across 16 aspects defined in VBench \citep{huang2024vbench}. We adopt DuCa’s norm‑based token ranking compatible with FlashAttention \citep{Dao2022FlashAttentionFA} for faster inference. DiffSparse achieves the highest VBench score while minimizing computational cost and inference time. At the same compression ratio, it delivers greater speedup by skipping partial layers with zero sparsity, and its adaptive, layer‑wise sparsity allocation preserves model quality.

\subsection{Ablation Studies}
\paragraph{Comparison of Two Stage Training.}
In this work, we adopt a two-stage training strategy. The first stage independently trains cost matrices for full-step and layer sparsity. In the second stage, the learned full-step cost is merged into the layer sparsity optimization, and the layer sparsity is subsequently fine-tuned. This design enables the model to initially leverage the full-step to correct errors and to learn layer sparsity cost values, followed by a gradual reduction of the full-step influence. Results proved that two-stage approach achieves better performance, with an FID of 26.91 compared to 27.40 from the single-stage baseline.

\paragraph{Comparison of Important Scores.}
\vspace{-8pt}
Table~\ref{table: Ablation-TokenSelection} compares three importance scores: attention (Equation \ref{eq:score}), cosine similarity and the $\ell_2$ norm. The cosine similarity is computed between the current input token and cached tokens. The $\ell_2$ norm is the norm value of input tokens.
The attention-based score attains the best FID, followed by the similarity measure, which captures token redundancy effectively. Norm-based scoring introduces noise and performs worst, confirming that accurate importance estimation is critical for optimal token selection.

\begin{center}
\vspace{-8pt}
\setlength{\tabcolsep}{4pt} 
\small

\begin{minipage}[t]{0.48\textwidth}
  \centering
    \captionof{table}{Ablation study on token importance metrics.}
  \label{table: Ablation-TokenSelection}
  \vspace{-8pt}
  \begin{tabular}{l|cc}
    \toprule
     Method & Base. & w/ DiffSparse \\
    \midrule
    Norm       & 29.05 & 28.89 \textcolor{red}{{(-0.16)}} \\
    Similarity & 29.00 & 28.07 \textcolor{red}{{(-0.93)}} \\
    Attention  & 28.35 & \textbf{26.91 \textcolor{red}{{(-1.44)}}} \\
    \bottomrule
  \end{tabular}
  \vspace{4pt}

\end{minipage}\hfill
\begin{minipage}[t]{0.48\textwidth}
  \centering
    \captionof{table}{Ablation study on distillation loss functions.}
  \label{tab:loss}
  \vspace{-8pt}
  \begin{tabular}{l|cc}
    \toprule
    Method & FID $\downarrow$ & CLIP $\uparrow$ \\
    \midrule
    L2    & 27.68 & 0.164 \\
    SSIM  & 27.46 & 0.164 \\
    LPIPS & \textbf{26.91} & \textbf{0.164} \\
    \bottomrule
  \end{tabular}
  \vspace{4pt}

\end{minipage}\hfill

\end{center}

\begin{center}
\vspace{-8pt}
\setlength{\tabcolsep}{4pt}
\small
\begin{minipage}[t]{0.48\textwidth}
  \centering
  \captionof{table}{{Ablation study of sparse interval.}}
  \label{table: Sparse-Interval}
  \vspace{-8pt}
  \begin{tabular}{l|c|cc}
    \toprule
     Interval & $|S|$ & FID $\downarrow$ & CLIP $\uparrow$ \\
    \midrule
    0.1   & 11 & 27.96 & 0.163 \\
    0.125 & 9 & 27.91 & 0.163 \\
    0.25  & 5 & \textbf{26.91} & 0.164 \\
    0.5   & 3 & 27.54 & \textbf{0.164} \\
    1.0   & 2  & 28.22 & 0.162 \\
    \bottomrule
  \end{tabular}
  \vspace{4pt}
\end{minipage}\hfill
\begin{minipage}[t]{0.48\textwidth}
  \centering
    \captionof{table}{{Ablation of warm-start strength \(\delta\).}}
    \label{tab:delta-ablation}
    \begin{tabular}{lcc}
    \toprule
    \(\delta\) & FID \(\downarrow\) & CLIP \(\uparrow\) \\
    \midrule
    0  & 27.40 & 0.163 \\
    5  & 27.01 & 0.164 \\
    10 & \textbf{26.91} & \textbf{0.164} \\
    20 & 26.95 & 0.164 \\
    \bottomrule
    \end{tabular}
  \vspace{4pt}
\end{minipage}

\end{center}

\paragraph{Comparison of Training Losses.}
\vspace{-8pt}
We compare L2, SSIM \citep{wang2004image}, and LPIPS losses in Table~\ref{tab:loss}. LPIPS outperforms the others, yielding the best FID. L2 loss penalizes pixel‐wise squared errors and often produces overly smooth images that lack fine details. SSIM enforces local structural similarity but may over‐penalize perceptually good images that differ spatially from the original. By measuring distances in a learned perceptual feature space, LPIPS avoids these pitfalls and better preserves image quality during training.

\paragraph{Comparison of Sparse Intervals.}
\vspace{-8pt}
We distribute sparsity uniformly across layers by token count and evaluate different granularity settings in Table~\ref{table: Sparse-Interval}. A granularity of 0.125 yields minimal within‑layer variation, which hinders convergence, while 0.5 limits the range of sparsity choices. The optimal granularity is 0.25, producing sparsity rates [0, 0.25, 0.50, 0.75, 1.0] (corresponding to candidate token counts of [0, 64, 128, 192, 256]  for a sequence length of 256) and delivering the best performance.

\vspace{-8pt}
\paragraph{Generalization on Higher Resolution Models.}
As the token sequence length increases with image resolution, peak memory usage during training grows substantially, even though the size and computational cost of our cost matrix remain unchanged. This makes direct training at very high resolutions impractical. To address this, we investigate whether a sparsity predictor trained at lower resolution can be {\textbf{transferred to higher resolution without retraining}}. As shown in Table \ref{tab:pixart-512}, the sparsity predictor learned at 256 × 256 resolution achieves a lower FID than ToCa on 512 × 512 images while maintaining a comparable CLIP-Score to the original PixArt model. These results demonstrate that our method generalizes effectively to higher resolutions, enabling model acceleration with limited memory and training cost.

\vspace{-8pt}
\paragraph{Compared with GA Search.}
We compared DiffSparse against traditional search methods such as random search and genetic algorithms and found that in the vast sparsity space they underperform. After 1,000 iterations on 500 images, these methods yield FID scores of 28.34 and 27.94, respectively, compared with 26.91 for DiffSparse. Moreover, they require about 16 hours, whereas DiffSparse completes training in roughly 4 hours. These results show that our differentiable learning framework discovers more effective layer-wise sparsity allocations and delivers superior acceleration.

\paragraph{{Comparison of Warm-Start Constant $\delta$.}}
{
Algorithm~\ref{algorithm:TwoStageTraining} uses a warm-start constant \(\delta=10\) for the two-stage optimization. Intuitively, \(\delta\) injects the Stage-1 prior (the timesteps selected to remain full-step) into Stage~2 by lowering the cost of the ``full'' candidate at those timesteps. In effect, a larger \(\delta\) more strongly encourages preserving full computation at the Stage-1 selected steps.
To quantify this effect we evaluated \(\delta\in\{0,5,10,20\}\). Table~\ref{tab:delta-ablation} reports the results on PixArt-\(\alpha\) with \(T=20\). A moderate warm-start (\(\delta=10\)) recovers most of the benefit, while \(\delta=0\) (no warm-start) removes the Stage-1 prior and yields noticeably worse performance.
}

\subsection{Qualitative Analysis}
\begin{figure}[h]
    \centering
    \includegraphics[width=0.8\linewidth]{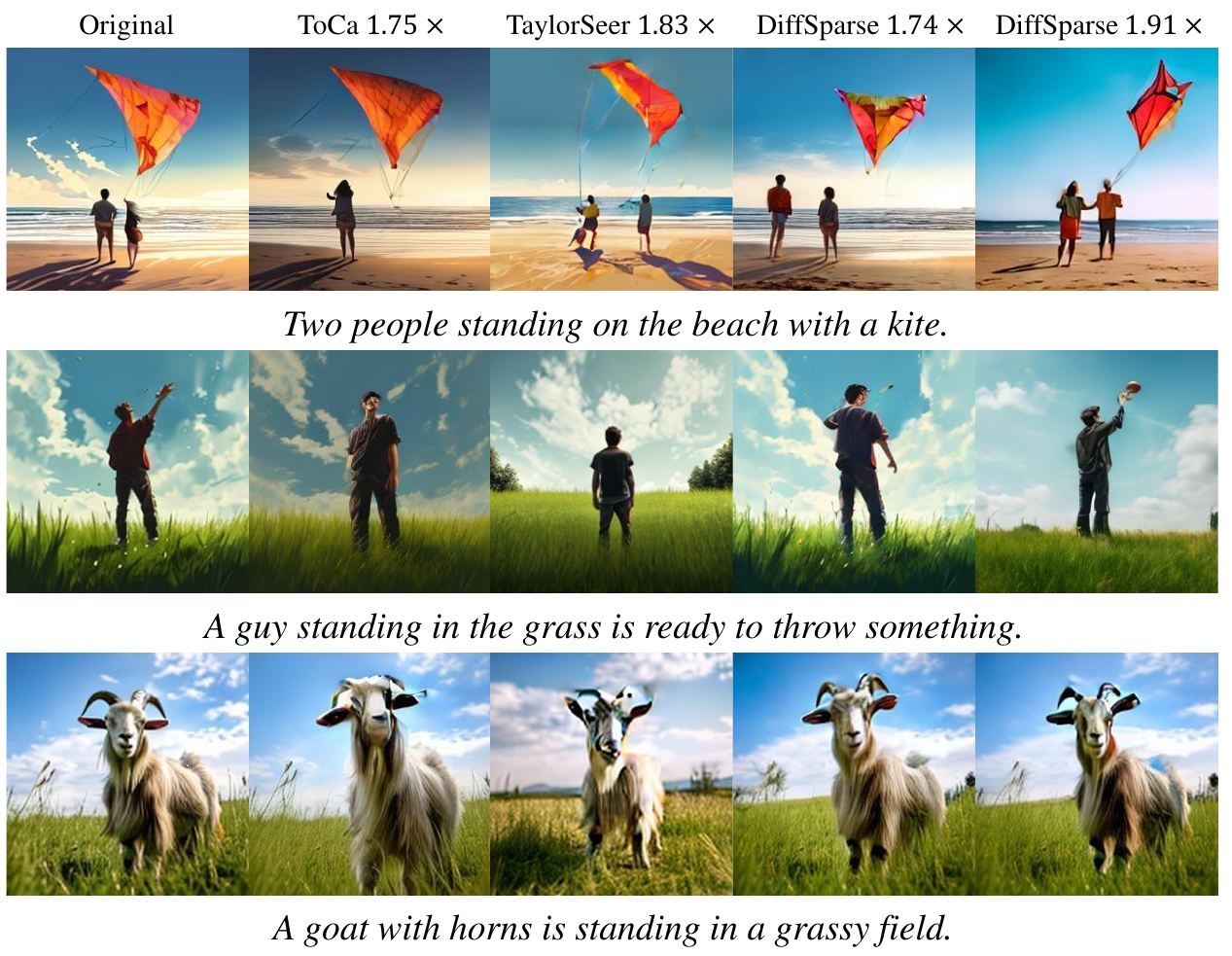}
    \caption{Comparison of our method with the baseline (PixArt-$\alpha$ with DPM-Solver++ using 20 steps) and existing methods under different acceleration rates.}
    \label{fig:pruning_comparison}
    \vspace{-10pt}
\end{figure}

\paragraph{Visualization of Generated Images.}
We provide detailed visual comparisons among our proposed method, ToCa, and the original PixArt-\(\alpha\) across various sparsity ratios. in Figure \ref{fig:pruning_comparison} reveals that DiffSparse consistently maintains high fidelity, even under aggressive pruning conditions. Moreover, our DiffSparse effectively preserves the semantic content of the text prompt, ensuring that the generated images remain closely aligned with the original descriptions. In contrast, baseline methods exhibit noticeable degradation in both visual quality and text-image alignment at higher pruning ratios, further highlighting the strength and efficiency of DiffSparse.

\vspace{-8pt}
\section{Conclusion}
We introduce a learnable token sparsity allocation framework to accelerate diffusion transformers. By formulating sparsity allocation as a dynamic programming problem and employing a two stage training strategy, our method substantially reduces computational cost while preserving generative quality. Extensive experiments across various foundation models and datasets demonstrate improved acceleration ratios without compromising image quality of our method.

\bibliography{main}

@article{bu2025dicache,
  title={Dicache: Let diffusion model determine its own cache},
  author={Bu, Jiazi and Ling, Pengyang and Zhou, Yujie and Wang, Yibin and Zang, Yuhang and Lin, Dahua and Wang, Jiaqi},
  journal={arXiv preprint arXiv:2508.17356},
  year={2025}
}

@article{zhu2026tap,
  title={TAP: A Token-Adaptive Predictor Framework for Training-Free Diffusion Acceleration},
  author={Zhu, Haowei and Huang, Tingxuan and Wang, Xing and Zhao, Tianyu and Wang, Jiexi and Chen, Weifeng and Peng, Xurui and Chen, Fangmin and Yong, Junhai and Wang, Bin},
  journal={arXiv preprint arXiv:2603.03792},
  year={2026}
}

@String(AAAI = {AAAI})

@article{structural_pruning_diffusion,
  title={Structural Pruning for Diffusion Models},
  author={Fang, Gongfan and Ma, Xinyin and Wang, Xinchao},
  journal={arXiv preprint arXiv:2305.10924},
  year={2023}
}

@article{selvaraju2024fora,
  title={Fora: Fast-forward caching in diffusion transformer acceleration},
  author={Selvaraju, Pratheba and Ding, Tianyu and Chen, Tianyi and Zharkov, Ilya and Liang, Luming},
  journal={arXiv preprint arXiv:2407.01425},
  year={2024}
}

@article{salimans2022progressive,
  title={Progressive distillation for fast sampling of diffusion models},
  author={Salimans, Tim and Ho, Jonathan},
  journal={arXiv preprint arXiv:2202.00512},
  year={2022}
}

@inproceedings{
zou2025accelerating,
title={Accelerating Diffusion Transformers with Token-wise Feature Caching},
author={Chang Zou and Xuyang Liu and Ting Liu and Siteng Huang and Linfeng Zhang},
booktitle={The Thirteenth International Conference on Learning Representations},
year={2025},
url={https://openreview.net/forum?id=yYZbZGo4ei}
}

@article{zhu2025dip,
  title={DiP-GO: A Diffusion Pruner via Few-step Gradient Optimization},
  author={Zhu, Haowei and Tang, Dehua and Liu, Ji and Lu, Mingjie and Zheng, Jintu and Peng, Jinzhang and Li, Dong and Wang, Yu and Jiang, Fan and Tian, Lu and others},
  journal={Advances in Neural Information Processing Systems},
  volume={37},
  pages={92581--92604},
  year={2025}
}

@article{vaswani2017attention,
  title={Attention is all you need},
  author={Vaswani, Ashish and Shazeer, Noam and Parmar, Niki and Uszkoreit, Jakob and Jones, Llion and Gomez, Aidan N and Kaiser, {\L}ukasz and Polosukhin, Illia},
  journal={Advances in neural information processing systems},
  volume={30},
  year={2017}
}

@inproceedings{ronneberger2015u,
  title={U-net: Convolutional networks for biomedical image segmentation},
  author={Ronneberger, Olaf and Fischer, Philipp and Brox, Thomas},
  booktitle={Medical image computing and computer-assisted intervention--MICCAI 2015: 18th international conference, Munich, Germany, October 5-9, 2015, proceedings, part III 18},
  pages={234--241},
  year={2015},
  organization={Springer}
}

@article{podell2023sdxl,
  title={Sdxl: Improving latent diffusion models for high-resolution image synthesis},
  author={Podell, Dustin and English, Zion and Lacey, Kyle and Blattmann, Andreas and Dockhorn, Tim and M{\"u}ller, Jonas and Penna, Joe and Rombach, Robin},
  journal={arXiv preprint arXiv:2307.01952},
  year={2023}
}

@inproceedings{rombach2022high,
  title={High-resolution image synthesis with latent diffusion models},
  author={Rombach, Robin and Blattmann, Andreas and Lorenz, Dominik and Esser, Patrick and Ommer, Bj{\"o}rn},
  booktitle={Proceedings of the IEEE/CVF conference on computer vision and pattern recognition},
  pages={10684--10695},
  year={2022}
}

@article{ho2020denoising,
  title={Denoising diffusion probabilistic models},
  author={Ho, Jonathan and Jain, Ajay and Abbeel, Pieter},
  journal={Advances in neural information processing systems},
  volume={33},
  pages={6840--6851},
  year={2020}
}

@inproceedings{fang2023structural,
  title={Structural pruning for diffusion models},
  author={Gongfan Fang and Xinyin Ma and Xinchao Wang},
  booktitle={Advances in Neural Information Processing Systems},
  year={2023},
}

@article{song2020denoising,
  title={Denoising diffusion implicit models},
  author={Song, Jiaming and Meng, Chenlin and Ermon, Stefano},
  journal={arXiv preprint arXiv:2010.02502},
  year={2020}
}

@misc{ho2022imagenvideohighdefinition,
      title={Imagen Video: High Definition Video Generation with Diffusion Models}, 
      author={Jonathan Ho and William Chan and Chitwan Saharia and Jay Whang and Ruiqi Gao and Alexey Gritsenko and Diederik P. Kingma and Ben Poole and Mohammad Norouzi and David J. Fleet and Tim Salimans},
      year={2022},
      eprint={2210.02303},
      archivePrefix={arXiv},
      primaryClass={cs.CV},
      url={https://arxiv.org/abs/2210.02303}, 
}

@article{saharia2022photorealistic,
  title={Photorealistic text-to-image diffusion models with deep language understanding},
  author={Saharia, Chitwan and Chan, William and Saxena, Saurabh and Li, Lala and Whang, Jay and Denton, Emily L and Ghasemipour, Kamyar and Gontijo Lopes, Raphael and Karagol Ayan, Burcu and Salimans, Tim and others},
  journal={Advances in neural information processing systems},
  volume={35},
  pages={36479--36494},
  year={2022}
}

@article{chen2023pixart,
  title={Pixart-${\alpha}$: Fast training of diffusion transformer for photorealistic text-to-image synthesis},
  author={Chen, Junsong and Yu, Jincheng and Ge, Chongjian and Yao, Lewei and Xie, Enze and Wu, Yue and Wang, Zhongdao and Kwok, James and Luo, Ping and Lu, Huchuan and others},
  journal={arXiv preprint arXiv:2310.00426},
  year={2023}
}

@inproceedings{peebles2023scalable,
  title={Scalable diffusion models with transformers},
  author={Peebles, William and Xie, Saining},
  booktitle={Proceedings of the IEEE/CVF international conference on computer vision},
  pages={4195--4205},
  year={2023}
}

@inproceedings{ma2024deepcache,
  title={Deepcache: Accelerating diffusion models for free},
  author={Ma, Xinyin and Fang, Gongfan and Wang, Xinchao},
  booktitle={Proceedings of the IEEE/CVF Conference on Computer Vision and Pattern Recognition},
  pages={15762--15772},
  year={2024}
}

@article{li2023FasterDiffusion,
  title={Faster diffusion: Rethinking the role of unet encoder in diffusion models},
  author={Li, Senmao and Hu, Taihang and Khan, Fahad Shahbaz and Li, Linxuan and Yang, Shiqi and Wang, Yaxing and Cheng, Ming-Ming and Yang, Jian},
  journal={arXiv preprint arXiv:2312.09608},
  year={2023}
}

@article{chen2024delta-dit,
  title={$\Delta$-DiT: A Training-Free Acceleration Method Tailored for Diffusion Transformers},
  author={Chen, Pengtao and Shen, Mingzhu and Ye, Peng and Cao, Jianjian and Tu, Chongjun and Bouganis, Christos-Savvas and Zhao, Yiren and Chen, Tao},
  journal={arXiv preprint arXiv:2406.01125},
  year={2024}
}

@inproceedings{wimbauer2024cache,
  title={Cache me if you can: Accelerating diffusion models through block caching},
  author={Wimbauer, Felix and Wu, Bichen and Schoenfeld, Edgar and Dai, Xiaoliang and Hou, Ji and He, Zijian and Sanakoyeu, Artsiom and Zhang, Peizhao and Tsai, Sam and Kohler, Jonas and others},
  booktitle={Proceedings of the IEEE/CVF Conference on Computer Vision and Pattern Recognition},
  pages={6211--6220},
  year={2024}
}

@inproceedings{songDDIM,
  title={Denoising Diffusion Implicit Models},
  author={Song, Jiaming and Meng, Chenlin and Ermon, Stefano},
  booktitle={International Conference on Learning Representations},
  year={2021}
}

@inproceedings{peebles2023dit,
  title={Scalable diffusion models with transformers},
  author={Peebles, William and Xie, Saining},
  booktitle={Proceedings of the IEEE/CVF International Conference on Computer Vision},
  pages={4195--4205},
  year={2023}
}

@inproceedings{chen2023pixartalpha,
      title={PixArt-${\alpha}$: Fast Training of Diffusion Transformer for Photorealistic Text-to-Image Synthesis}, 
      author={Junsong Chen and Jincheng Yu and Chongjian Ge and Lewei Yao and Enze Xie and Yue Wu and Zhongdao Wang and James Kwok and Ping Luo and Huchuan Lu and Zhenguo Li},
      booktitle={International Conference on Learning Representations},
      year={2024}
}

@article{openai2024sora,
  title={Video generation models as world simulators},
  author={Tim Brooks and Bill Peebles and Connor Holmes and Will DePue and Yufei Guo and Li Jing and David Schnurr and Joe Taylor and Troy Luhman and Eric Luhman and Clarence Ng and Ricky Wang and Aditya Ramesh},
  year={2024},
  url={https://openai.com/research/video-generation-models-as-world-simulators},
}

@article{duca,
  title={Accelerating Diffusion Transformers with Dual Feature Caching},
  author={Zou, Chang and Zhang, Evelyn and Guo, Runlin and Xu, Haohang and He, Conghui and Hu, Xuming and Zhang, Linfeng},
  journal={arXiv preprint arXiv:2412.18911},
  year={2024}
}

@article{lu2022dpm,
  title={Dpm-solver: A fast ode solver for diffusion probabilistic model sampling in around 10 steps},
  author={Lu, Cheng and Zhou, Yuhao and Bao, Fan and Chen, Jianfei and Li, Chongxuan and Zhu, Jun},
  journal={Advances in Neural Information Processing Systems},
  volume={35},
  pages={5775--5787},
  year={2022}
}

@article{lu2022dpm++,
  title={Dpm-solver++: Fast solver for guided sampling of diffusion probabilistic models},
  author={Lu, Cheng and Zhou, Yuhao and Bao, Fan and Chen, Jianfei and Li, Chongxuan and Zhu, Jun},
  journal={arXiv preprint arXiv:2211.01095},
  year={2022}
}

@article{Dao2022FlashAttentionFA,
  title={FlashAttention: Fast and Memory-Efficient Exact Attention with IO-Awareness},
  author={Tri Dao and Daniel Y. Fu and Stefano Ermon and Atri Rudra and Christopher R'e},
  journal={ArXiv},
  year={2022},
  volume={abs/2205.14135},
  url={https://api.semanticscholar.org/CorpusID:249151871}
}

@article{zhang2024laptop,
  title={Laptop-diff: Layer pruning and normalized distillation for compressing diffusion models},
  author={Zhang, Dingkun and Li, Sijia and Chen, Chen and Xie, Qingsong and Lu, Haonan},
  journal={arXiv preprint arXiv:2404.11098},
  year={2024}
}

@inproceedings{yin2024stepdistillation,
  title={One-step diffusion with distribution matching distillation},
  author={Yin, Tianwei and Gharbi, Micha{\"e}l and Zhang, Richard and Shechtman, Eli and Durand, Fredo and Freeman, William T and Park, Taesung},
  booktitle={Proceedings of the IEEE/CVF Conference on Computer Vision and Pattern Recognition},
  pages={6613--6623},
  year={2024}
}

@article{luo2023latent,
  title={Latent consistency models: Synthesizing high-resolution images with few-step inference},
  author={Luo, Simian and Tan, Yiqin and Huang, Longbo and Li, Jian and Zhao, Hang},
  journal={arXiv preprint arXiv:2310.04378},
  year={2023}
}

@inproceedings{deng2009imagenet,
  title={Imagenet: A large-scale hierarchical image database},
  author={Deng, Jia and Dong, Wei and Socher, Richard and Li, Li-Jia and Li, Kai and Fei-Fei, Li},
  booktitle={2009 IEEE conference on computer vision and pattern recognition},
  pages={248--255},
  year={2009},
  organization={Ieee}
}

@inproceedings{huang2024vbench,
  title={Vbench: Comprehensive benchmark suite for video generative models},
  author={Huang, Ziqi and He, Yinan and Yu, Jiashuo and Zhang, Fan and Si, Chenyang and Jiang, Yuming and Zhang, Yuanhan and Wu, Tianxing and Jin, Qingyang and Chanpaisit, Nattapol and others},
  booktitle={Proceedings of the IEEE/CVF Conference on Computer Vision and Pattern Recognition},
  pages={21807--21818},
  year={2024}
}

@article{heusel2017fid,
  title={Gans trained by a two time-scale update rule converge to a local nash equilibrium},
  author={Heusel, Martin and Ramsauer, Hubert and Unterthiner, Thomas and Nessler, Bernhard and Hochreiter, Sepp},
  journal={Advances in neural information processing systems},
  volume={30},
  year={2017}
}

@inproceedings{lin2014coco,
  title={Microsoft coco: Common objects in context},
  author={Lin, Tsung-Yi and Maire, Michael and Belongie, Serge and Hays, James and Perona, Pietro and Ramanan, Deva and Doll{\'a}r, Piotr and Zitnick, C Lawrence},
  booktitle={Computer Vision--ECCV 2014: 13th European Conference, Zurich, Switzerland, September 6-12, 2014, Proceedings, Part V 13},
  pages={740--755},
  year={2014},
  organization={Springer}
}

@article{hessel2021clipscore,
  title={Clipscore: A reference-free evaluation metric for image captioning},
  author={Hessel, Jack and Holtzman, Ari and Forbes, Maxwell and Bras, Ronan Le and Choi, Yejin},
  journal={arXiv preprint arXiv:2104.08718},
  year={2021}
}

@article{wang2004image,
  title={Image quality assessment: from error visibility to structural similarity},
  author={Wang, Zhou and Bovik, Alan C and Sheikh, Hamid R and Simoncelli, Eero P},
  journal={IEEE transactions on image processing},
  volume={13},
  number={4},
  pages={600--612},
  year={2004},
  publisher={IEEE}
}

@article{zhang2018lpips,
      title={The Unreasonable Effectiveness of Deep Features as a Perceptual Metric}, 
      author={Richard Zhang and Phillip Isola and Alexei A. Efros and Eli Shechtman and Oliver Wang},
      year={2018},
      eprint={1801.03924},
      archivePrefix={arXiv},
      primaryClass={cs.CV},
      url={https://arxiv.org/abs/1801.03924}, 
}

@article{jang2016categorical,
  title={Categorical reparameterization with gumbel-softmax},
  author={Jang, Eric and Gu, Shixiang and Poole, Ben},
  journal={arXiv preprint arXiv:1611.01144},
  year={2016}
}

@inproceedings{esser2024scaling,
  title={Scaling rectified flow transformers for high-resolution image synthesis},
  author={Esser, Patrick and Kulal, Sumith and Blattmann, Andreas and Entezari, Rahim and M{\"u}ller, Jonas and Saini, Harry and Levi, Yam and Lorenz, Dominik and Sauer, Axel and Boesel, Frederic and others},
  booktitle={Forty-first international conference on machine learning},
  year={2024}
}

@article{tian2024u,
  title={U-dits: Downsample tokens in u-shaped diffusion transformers},
  author={Tian, Yuchuan and Tu, Zhijun and Chen, Hanting and Hu, Jie and Xu, Chao and Wang, Yunhe},
  journal={arXiv preprint arXiv:2405.02730},
  year={2024}
}

@article{chen2024pixart,
title={PIXART-${\delta}$: Fast and Controllable Image Generation with Latent Consistency Models}, 
      author={Junsong Chen and Yue Wu and Simian Luo and Enze Xie and Sayak Paul and Ping Luo and Hang Zhao and Zhenguo Li},
      year={2024},
      eprint={2401.05252},
      archivePrefix={arXiv},
      primaryClass={cs.CV}
}

@misc{flux2024,
    author={Black Forest Labs},
    title={FLUX},
    year={2024},
    howpublished={\url{https://github.com/black-forest-labs/flux}},
}

@inproceedings{xu2023imagereward,
  title={ImageReward: learning and evaluating human preferences for text-to-image generation},
  author={Xu, Jiazheng and Liu, Xiao and Wu, Yuchen and Tong, Yuxuan and Li, Qinkai and Ding, Ming and Tang, Jie and Dong, Yuxiao},
  booktitle={Proceedings of the 37th International Conference on Neural Information Processing Systems},
  pages={15903--15935},
  year={2023}
}

@article{yu2022scaling,
  title={Scaling autoregressive models for content-rich text-to-image generation},
  author={Yu, Jiahui and Xu, Yuanzhong and Koh, Jing Yu and Luong, Thang and Baid, Gunjan and Wang, Zirui and Vasudevan, Vijay and Ku, Alexander and Yang, Yinfei and Ayan, Burcu Karagol and others},
  journal={arXiv preprint arXiv:2206.10789},
  volume={2},
  number={3},
  pages={5},
  year={2022}
}

@article{lyu2020deepfake,
  title={Deepfake detection: Current challenges and next steps},
  author={Lyu, Siwei},
  booktitle={ICMEW},
  pages={1--6},
  year={2020},
  organization={IEEE}
}

@article{wan2025wan,
  title={Wan: Open and advanced large-scale video generative models},
  author={Wan, Team and Wang, Ang and Ai, Baole and Wen, Bin and Mao, Chaojie and Xie, Chen-Wei and Chen, Di and Yu, Feiwu and Zhao, Haiming and Yang, Jianxiao and others},
  journal={arXiv preprint arXiv:2503.20314},
  year={2025}
}

@inproceedings{zhang2025training,
  title={Training-free and hardware-friendly acceleration for diffusion models via similarity-based token pruning},
  author={Zhang, Evelyn and Tang, Jiayi and Ning, Xuefei and Zhang, Linfeng},
  booktitle={Proceedings of the AAAI Conference on Artificial Intelligence},
  volume={39},
  number={9},
  pages={9878--9886},
  year={2025}
}

@inproceedings{you2025layer,
  title={Layer-and Timestep-Adaptive Differentiable Token Compression Ratios for Efficient Diffusion Transformers},
  author={You, Haoran and Barnes, Connelly and Zhou, Yuqian and Kang, Yan and Du, Zhenbang and Zhou, Wei and Zhang, Lingzhi and Nitzan, Yotam and Liu, Xiaoyang and Lin, Zhe and others},
  booktitle={Proceedings of the Computer Vision and Pattern Recognition Conference},
  pages={18072--18082},
  year={2025}
}

@article{liu2022flow,
  title={Flow straight and fast: Learning to generate and transfer data with rectified flow},
  author={Liu, Xingchao and Gong, Chengyue and Liu, Qiang},
  journal={arXiv preprint arXiv:2209.03003},
  year={2022}
}

@article{liu2025reusing,
  title={From reusing to forecasting: Accelerating diffusion models with taylorseers},
  author={Liu, Jiacheng and Zou, Chang and Lyu, Yuanhuiyi and Chen, Junjie and Zhang, Linfeng},
  journal={arXiv preprint arXiv:2503.06923},
  year={2025}
}

@inproceedings{liu2025timestep,
  title={Timestep Embedding Tells: It's Time to Cache for Video Diffusion Model},
  author={Liu, Feng and Zhang, Shiwei and Wang, Xiaofeng and Wei, Yujie and Qiu, Haonan and Zhao, Yuzhong and Zhang, Yingya and Ye, Qixiang and Wan, Fang},
  booktitle={Proceedings of the Computer Vision and Pattern Recognition Conference},
  pages={7353--7363},
  year={2025}
}

@InProceedings{Bain21,
  author       = "Max Bain and Arsha Nagrani and G{\"u}l Varol and Andrew Zisserman",
  title        = "Frozen in Time: A Joint Video and Image Encoder for End-to-End Retrieval",
  booktitle    = "IEEE International Conference on Computer Vision",
  year         = "2021",
}

@article{yuan2024ditfastattn,
  title={Ditfastattn: Attention compression for diffusion transformer models},
  author={Yuan, Zhihang and Zhang, Hanling and Pu, Lu and Ning, Xuefei and Zhang, Linfeng and Zhao, Tianchen and Yan, Shengen and Dai, Guohao and Wang, Yu},
  journal={Advances in Neural Information Processing Systems},
  volume={37},
  pages={1196--1219},
  year={2024}
}

@article{qiu2025accelerating,
  title={Accelerating Diffusion Transformer via Error-Optimized Cache},
  author={Qiu, Junxiang and Wang, Shuo and Lu, Jinda and Liu, Lin and Jiang, Houcheng and Zhu, Xingyu and Hao, Yanbin},
  journal={arXiv preprint arXiv:2501.19243},
  year={2025}
}

@article{sun2025unicp,
  title={UniCP: A Unified Caching and Pruning Framework for Efficient Video Generation},
  author={Sun, Wenzhang and Hou, Qirui and Di, Donglin and Yang, Jiahui and Ma, Yongjia and Cui, Jianxun},
  journal={arXiv preprint arXiv:2502.04393},
  year={2025}
}

@article{liu2025region,
  title={Region-adaptive sampling for diffusion transformers},
  author={Liu, Ziming and Yang, Yifan and Zhang, Chengruidong and Zhang, Yiqi and Qiu, Lili and You, Yang and Yang, Yuqing},
  journal={arXiv preprint arXiv:2502.10389},
  year={2025}
}

@article{liu2025speca,
  title={SpeCa: Accelerating Diffusion Transformers with Speculative Feature Caching},
  author={Liu, Jiacheng and Zou, Chang and Lyu, Yuanhuiyi and Ren, Fei and Wang, Shaobo and Li, Kaixin and Zhang, Linfeng},
  journal={arXiv preprint arXiv:2509.11628},
  year={2025}
}

@inproceedings{yin2024improved,
    title={Improved Distribution Matching Distillation for Fast Image Synthesis},
    author={Yin, Tianwei and Gharbi, Micha{\"e}l and Park, Taesung and Zhang, Richard and Shechtman, Eli and Durand, Fredo and Freeman, William T},
    booktitle={NeurIPS},
    year={2024}
}

@article{zhao2024dynamic,
  title={Dynamic diffusion transformer},
  author={Zhao, Wangbo and Han, Yizeng and Tang, Jiasheng and Wang, Kai and Song, Yibing and Huang, Gao and Wang, Fan and You, Yang},
  journal={arXiv preprint arXiv:2410.03456},
  year={2024}
}

@article{zhu2024distribution,
  title={Distribution-aware data expansion with diffusion models},
  author={Zhu, Haowei and Yang, Ling and Yong, Jun-Hai and Yin, Hongzhi and Jiang, Jiawei and Xiao, Meng and Zhang, Wentao and Wang, Bin},
  journal={Advances in Neural Information Processing Systems},
  volume={37},
  pages={102768--102795},
  year={2024}
}

@article{zhu2025recon,
  title={ReCon: Region-Controllable Data Augmentation with Rectification and Alignment for Object Detection},
  author={Zhu, Haowei and Pan, Tianxiang and Qin, Rui and Yong, Jun-Hai and Wang, Bin},
  journal={arXiv preprint arXiv:2510.15783},
  year={2025}
}

@inproceedings{wu2025drivescape,
  title={DriveScape: High-Resolution Driving Video Generation by Multi-View Feature Fusion},
  author={Wu, Wei and Guo, Xi and Tang, Weixuan and Huang, Tingxuan and Wang, Chiyu and Ding, Chenjing},
  booktitle={Proceedings of the Computer Vision and Pattern Recognition Conference},
  pages={17187--17196},
  year={2025}
}
\bibliographystyle{iclr2026_conference}

\appendix

\newpage

\section{Appendix}

\subsection{Ethical Statement}
\label{appendix:social_impact}
Generative models have shown impressive capabilities in content creation \citep{chen2023pixart, rombach2022high}, but their high inference costs hinder rapid deployment. Our method offers an efficient acceleration strategy for diffusion models, achieving near-lossless speedup without retraining and maintaining compatibility with various architectures. This generalizability makes it well-suited for fast deployment on mobile and edge devices.

However, generative models pretrained on large-scale internet data may reflect inherent social biases and stereotypes. There is also potential for misuse, such as in DeepFake \citep{lyu2020deepfake} creation, which can cause serious societal harm. As the cost of generation decreases, the risk of irresponsible use increases. Therefore, it's essential to establish regulations, foster a well-governed community, and provide clear usage guidelines to ensure the responsible application of generative technologies.

\subsection{Reproducibility Statement}
To support reproducibility, we provide detailed pseudocode for the proposed method (Appendix \ref{appendix:two_stage_training}), full training and evaluation protocols including all hyperparameters and optimizer settings (Section \ref{sec:experments_setting_v2} and Appendix \ref{appendix:more_detail}), dataset descriptions and preprocessing steps, and the computing environment for experiments (Section \ref{sec:experments_setting_v2}). Where applicable, we report evaluation metrics and include instructions sufficient to reproduce the experimental pipelines described in the main text and appendices. We plan to release anonymized code, trained model checkpoints, and exact run scripts upon paper acceptance to facilitate full replication.

\subsection{The Use of Large Language Models (LLMs)}
We did not rely on LLMs for research ideation, experiment design, data analysis, or the generation of technical content. Any use of LLMs was limited to minor, general-purpose editorial assistance (proofreading, grammar, phrasing, and formatting suggestions); all such suggestions were reviewed and revised by the authors. The paper’s conceptual contributions, algorithms, experiments, results, and conclusions were produced solely by the authors, and no LLM is credited as a contributor.

\subsection{{More Discussion with Existing Works}}
\subsubsection{{More Related Works}}

\paragraph{Diffusion Transformer Models.}
Recent work has improved the efficiency and scalability of transformer-based diffusion models. Hybrid CNN–transformer architectures~\citep{saharia2022photorealistic} combine local inductive biases with global attention, and transformer-based video generation~\citep{ho2022imagenvideohighdefinition} demonstrates strong temporal modeling. These results establish transformers as a versatile backbone for diffusion, motivating efforts on optimization, faster inference, and stronger conditional generation. Nevertheless, the iterative denoising loop still incurs substantial computational overhead that limits industrial deployment.

\paragraph{Acceleration of Diffusion Models.}
Several recent methods target inference cost directly: EOC leverages prior knowledge to improve caching~\citep{qiu2025accelerating}, while designs such as UniCP and RAS further boost efficiency~\citep{sun2025unicp,liu2025region}. {DyDiT~\citep{zhao2024dynamic} accelerates inference by skipping unimportant tokens and slimming per-layer width, whereas DiffSparse reduces compute by \emph{reusing} cached features. The two strategies are complementary and can be combined for larger speedups. DiffSparse computes token importance with a training-free compositional-attention score and learns a compact layer-wise predictor, leading to much faster convergence (on the order of \(10^{3}\) iterations versus DyDiT's \(\sim2\times10^{5}\) fine-tuning steps). Moreover, by optimizing a global \(T\)-step objective with a dynamic-programming solver, DiffSparse coordinates sparsity across timesteps and layers and is validated across multiple architectures and generation tasks.
}

\paragraph{{Comparison with Search-based and Training-based Methods.}}
{
In our main configurations, the differentiable optimization requires \(\approx 4\) hours of training versus \(\approx 16\) hours for a genetic-algorithm search baseline, DiffSparse attains better FID while using less optimization time. The learned sparsity predictor is compact (size \((T\times L)\times|S|\)) and often transfers from \(256\times256\) training to \(512\times512\) evaluation, reducing the need for retraining at higher resolutions. By contrast, distillation-based pipelines can demand orders of magnitude more compute: reported distillation efforts (DMD2) involve \(\mathcal{O}(10^{3}\text{--}10^{4})\) GPU·hours (e.g., SD1.5: 1,664 GPU·hr; SDXL: 8,192 GPU·hr), far exceeding the cost of our method and frequently relying on private data. Importantly, DiffSparse also improves some distilled models: for example, on a 4-step distilled model (FLUX.1-schnell) we observe a \(1.81\times\) speedup with no measurable quality drop (see Table~\ref{tab:flux}).
}

\subsection{More Implementation Details}
\label{appendix:more_detail}
\subsubsection{Token Selector}
\label{appendix:token_selector}
We rank tokens using a composite importance score that integrates four criteria: self-attention influence, cross-attention influence, cache reuse frequency, and uniform spatial distribution. This composite score, which has demonstrated effective in prior work~\citep{zou2025accelerating}, is defined for each token \(\hat{x}_i\) as follows: 
\begin{equation}
S(\hat{x}_i) = \mathcal{B}\Bigl( \lambda_1\, s_1(\hat{x}_i) + \lambda_2\, s_2(\hat{x}_i) + \lambda_3\, s_3(\hat{x}_i) \Bigr),
\label{eq:score}
\end{equation}

where \(s_1(\hat{x}_i) = \sum_{j=1}^{N} \alpha_{ij}\) quantifies the self-attention contribution of token \(\hat{x}_i\), with \(\alpha_{ij}\) being the \((i,j)\)-th element of the normalized self-attention matrix. A higher value indicates that the token exerts significant influence on others, meaning error in its representation may easily propagate. The term \(s_2(\hat{x}_i) = -\sum_{j=1}^{N} o_{ij}\log(o_{ij})\) represents the entropy of the cross-attention weights \(o_{ij}\), measuring how the control signal influences token \(\hat{x}_i\), with lower entropy indicating more focused guidance. Additionally, \(s_3(\hat{x}_i) = n_i\) denotes the number of times token \(\hat{x}_i\) has been reused from the cache since its last computation, where a higher \(n_i\) suggests possible accumulated errors, thus necessitating a fresh computation. The spatial bonus function \(\mathcal{B}(\cdot)\) promotes a uniform spatial distribution of the selected tokens by adding a bonus value \(\lambda_4\) to the score of \(\hat{x}_i\) if it has the highest composite score within its \(k \times k\) neighborhood. For each layer, tokens are ranked in descending order based on \(S(\hat{x}_i)\), and the top \(K\) tokens are selected for computation and cache updates according to a predefined sparsity ratio \(R\).

We adopt the hyperparameter settings recommended by ToCa \citep{zou2025accelerating} and DuCa \cite{duca} (which were shown to be optimal for that setup) and therefore do not include ablation experiments for these parameters, since tuning them is not central to our contribution.
Specifically, for PixArt-\(\alpha\), we set \(\lambda_1 = 0.0\), \(\lambda_2 = 1.0\), $\lambda_3 = 0.25/3$, \(\lambda_4 = 0.4\), and \(k = 4\).  For DiT, we use \(\lambda_1 = 1.0\), \(\lambda_2 = 0.0\), $\lambda_3 = 0.25/3$, \(\lambda_4 = 0.6\), and \(k = 2\). 
For FLUX.1-schnell, we set \(\lambda_1 = 0.0\), \(\lambda_2 = 1.0\), $\lambda_3 = 0.25/3$, \(\lambda_4 = 0.4\), and \(k = 4\). 

Besides, for Wan2.1, we select tokens with smaller norms in their value matrix as substitutes for those with high attention map scores, following a strategy shown to be effective in DuCa~\cite{duca}. Notably, our method does not introduce a new token-selector. Instead, it can be applied to existing token-selection methods and uses a differentiable sparsity-cost matrix to assign the model an optimal sparsity level. As shown in Table \ref{table: Ablation-TokenSelection}, across various token-importance metrics our approach consistently yields substantial gains.

\subsubsection{Layer Sparsity Cost Predictor}
We define a sparsity router for each layer. For PixArt-$\alpha$ and Wan2.1 model, each transformer block consists of a self-attention layer, a cross-attention layer, and an MLP layer, with each layer being assigned an individual sparsity value. In contrast, the DiT model does not include a cross-attention layer, thus the corresponding predictor for cross-attention layer is removed. Additionally, the FLUX model contains an image MLP layer, a text MLP layer, and a standard MLP layer, each of which is assigned its own sparsity value.

\subsubsection{Two-Stage Training}
\label{appendix:two_stage_training}
We present the pseudocode for our two-stage training algorithm in Algorithm \ref{algorithm:TwoStageTraining}, illustrating the training details of our sparsity cost predictor.

\begin{algorithm}[ht]
	\caption{Two-Stage Training Strategy for Cost Matrix Optimization}
	\label{algorithm:TwoStageTraining}
	\begin{algorithmic}
		\STATE {\bfseries Input:} Step cost matrix \(C_f \in \mathbb{R}^{T \times 2}\); Layer sparsity cost matrix \(C_l \in \mathbb{R}^{(L \times T) \times |S|}\); Total steps \(T\); Number of layers \(L\); Candidate set \(S\); Desired full-step count $|T_f|$; Mutation constant \(\delta = 10\).
		\STATE {\bfseries Stage 1: Initialization and Warm-Starting}
		\STATE Solve \(C_f\) via dynamic programming to obtain optimal full-step set $T_f$.
            \STATE Integrate \(C_f\) into the tuned \(C_l\) to form the unified cost matrix:
		\FOR{each \(t \in T_f\)}
			\FOR{\(l = 1\) to \(L\)}
				\STATE Update cost: \(C_l^{(t,l,N)} \gets C_l^{(t,l,N)} - \delta\).
			\ENDFOR
		\ENDFOR
		\STATE {\bfseries Stage 2: Unified Cost Optimization}
		
		\STATE Fine-tune the integrated \(C_l\) using differentiable cost interactions to systematically redistribute FLOPs across sampling steps.
		\STATE {\bfseries Output:} Optimized layer cost matrix \(C_l\).
	\end{algorithmic}
\end{algorithm}

\subsubsection{Search-based Approaches}
In this paper, we compare our method with search-based approaches. For the GA algorithm, we start by initializing a population of 50 $(T*L)$ layer-sparsity vectors that satisfy the sparsity requirements. Each candidate is evaluated using its FID value, which serves as the fitness score. In subsequent iterations, we select the best-performing individuals as parents for crossover operations and introduce mutations with a probability of 0.01 to maintain population diversity. This iterative process continues until the individual with the highest fitness score is identified. Besides, the random search algorithm generates a population of candidates that meet the sparsity requirements in a completely random manner. Their fitness is also evaluated using the FID value, and the optimal solution is updated iteratively until the maximum number of iterations is reached.

\subsubsection{{About the Retraining Requirement}}
{
If you change the \emph{model architecture} significantly (e.g., different number of layers $L$ or a different block structure), retraining or at least fine-tuning is required when the \emph{temporal} or \emph{architectural} axes change ($T$ or $L$) because its parameters are tied to $(T,L,|S|)$, but not usually when only token length (image resolution) increases. Given the modest one-time training cost (4--10 GPU-hours in our experiments) and the measurable quality, speed improvements, we believe the overhead is justified for deployed models where inference cost matters.
}

\subsection{More Experiments}
\label{appendix:more-exp}

\subsubsection{{Comparison on Distilled Model}}
{Feature caching leverages redundancy across timesteps but provides little benefit for distilled diffusion models with only one or two steps.} Nevertheless, we still evaluate DiffSparse on FLUX.1‐schnell \citep{flux2024} with 4 steps at 256×256 resolution on the PartiPrompts \citep{yu2022scaling} dataset. As Table~\ref{tab:flux} shows, DiffSparse attains the same acceleration rate as ToCa but yields a higher Image Reward, confirming its effectiveness in reallocating computation to the most critical layers and delivering lossless speedup. 

\begin{table}[h]
  \caption{Comparison in text-to-image generation for FLUX.1-schnell on PartiPrompts.}
  \small
  \label{tab:flux}
  \centering
  \begin{tabular}{l|cc}
    \toprule
    Method               & MACs (T)$\downarrow$ & Image Reward$\uparrow$  \\
    \midrule
    FLUX.1-schnell       & 13.247   & 1.064        \\ 
    75\% Steps           & 9.936    & 1.063        \\
    ToCa                 & 7.313    & 1.063        \\
    \rowcolor{gray!20}
        \textbf{DiffSparse }           & 7.316    & \textbf{1.184} \\
    \bottomrule
  \end{tabular}%

  \vspace{-10pt}
\end{table}

\subsubsection{{Ablations of the Attention-based Score}}
{Table~\ref{table: Ablation-TokenSelection} compares three importance metrics (attention-based score, cosine similarity, $\ell_2$ norm) and shows the attention-based score performs best overall. 
In Table \ref{tab:ablation_attention_score_components}, we further remove one component at a time ($-s_1$, $-s_2$, $-s_3$, $-B$) to answer how much each term contributes independently.}

\begin{table}[h]
  \caption{{Ablations of the attention-based score in text-to-image generation for PixArt-$\alpha$.}}
  \small
  \label{tab:ablation_attention_score_components}
  \centering
  \begin{tabular}{l|cc}
    \toprule
    Variant & FID$\downarrow$ & CLIP$\uparrow$ \\
    \midrule
    DiffSparse & \textbf{26.91} & \textbf{0.164} \\
    $-s_1$ (self-attention influence) & 27.11 & 0.164  \\
    $-s_2$ (cross-attention focus) & 27.48 & 0.163 \\
    $-s_3$ (cache-reuse frequency) & 27.23 & 0.164 \\
    $-B$ (spatial bonus) & 27.05 & 0.164  \\
    \bottomrule
  \end{tabular}%
\end{table}

\subsection{Qualitative Analysis}
\label{appendix:qualitative}
\subsubsection{Visualization of Layer Sparsity}
\begin{figure*}[htbp]
\centering
\begin{minipage}[b]{0.3\textwidth}
    \centering
    \includegraphics[width=\textwidth]{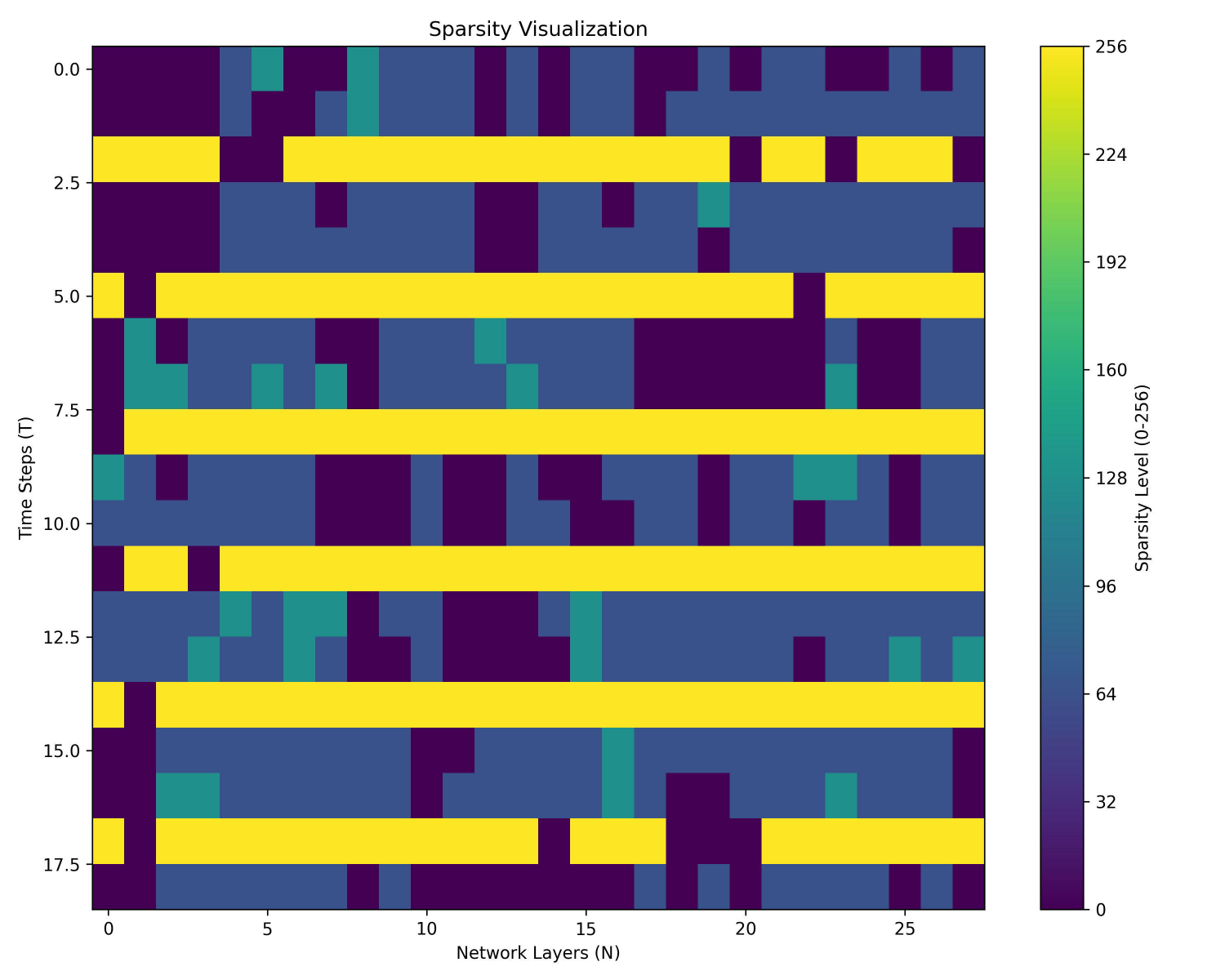}
    \small{(a) Self Attention layer.}
\end{minipage}
\begin{minipage}[b]{0.3\textwidth}
    \centering
    \includegraphics[width=\textwidth]{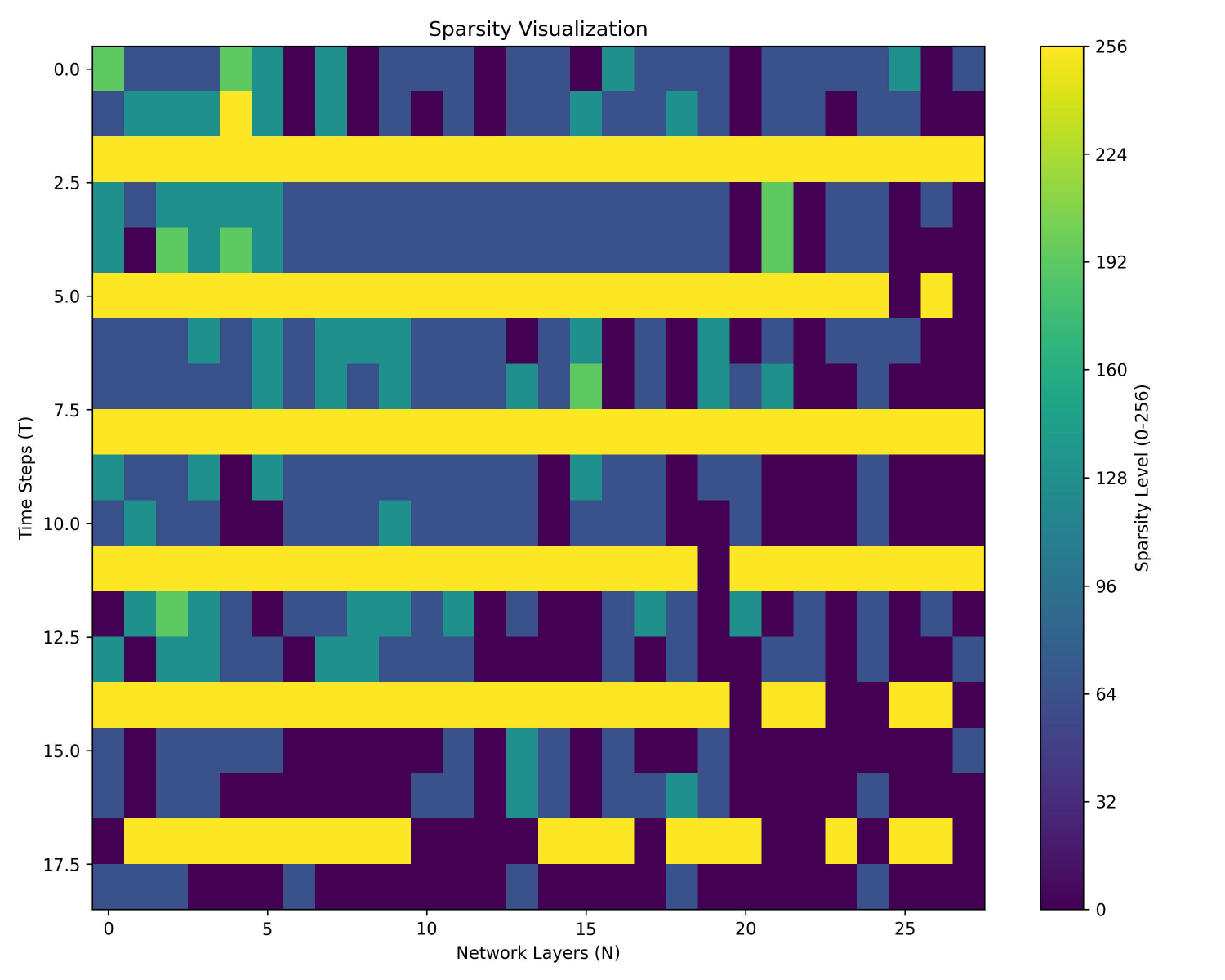}
     \small{(b) Cross Attention layer.}
\end{minipage}
\begin{minipage}[b]{0.3\textwidth}
    \centering
    \includegraphics[width=\textwidth]{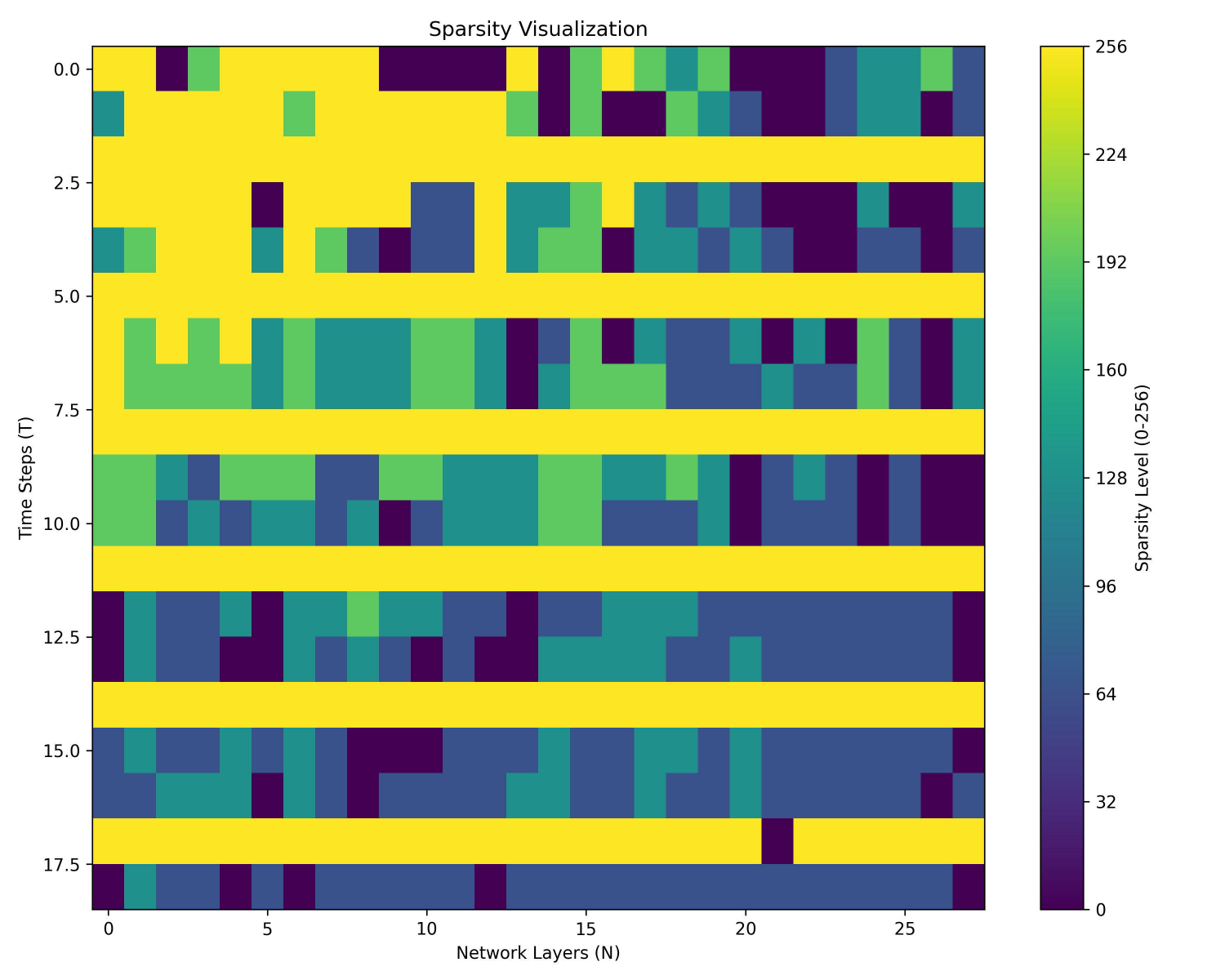}
     \small{(c) MLP layer.}
\end{minipage}
\caption{Visualization of predicted layer sparsity of PixArt-$\alpha$ with 20 steps. In the figure, the x-axis denotes different network layers, the y-axis denotes sampling time steps, and the color gradient from blue to yellow indicates increasing sparsity.}
\label{fig:layer_sparsity}
\end{figure*}

Figure \ref{fig:layer_sparsity} shows the learned layer wise sparsity allocation for PixArt-$\alpha$ at 256×256 resolution with 20 sampling steps under 1.74$\times$ speedup, with the first step omitted because it is always fully computed in cache-based acceleration methods. In the self attention layers, sparsity is higher (that is, the layers are more cacheable) in early time steps and shallow layers, while in the cross attention layers sparsity is lower in later time steps and deeper layers, suggesting that textual semantics are most important in the initial layers. The MLP layers receive more computation in early steps and shallow layers, with reduced sparsity in deep layers at early steps and in shallow layers at later steps. In addition, Figure \ref{fig:layer_sparsity} demonstrates that our method can redistribute the computation across all steps, reducing the dependence on fully computed steps. It shows that additional resources are allocated to MLP layer, this might be attributed to its ability of correct errors introduced by caching.

\subsubsection{More Visualization of Generated Images} 
\begin{figure}[h]
    \centering
    \includegraphics[width=1.0\linewidth]{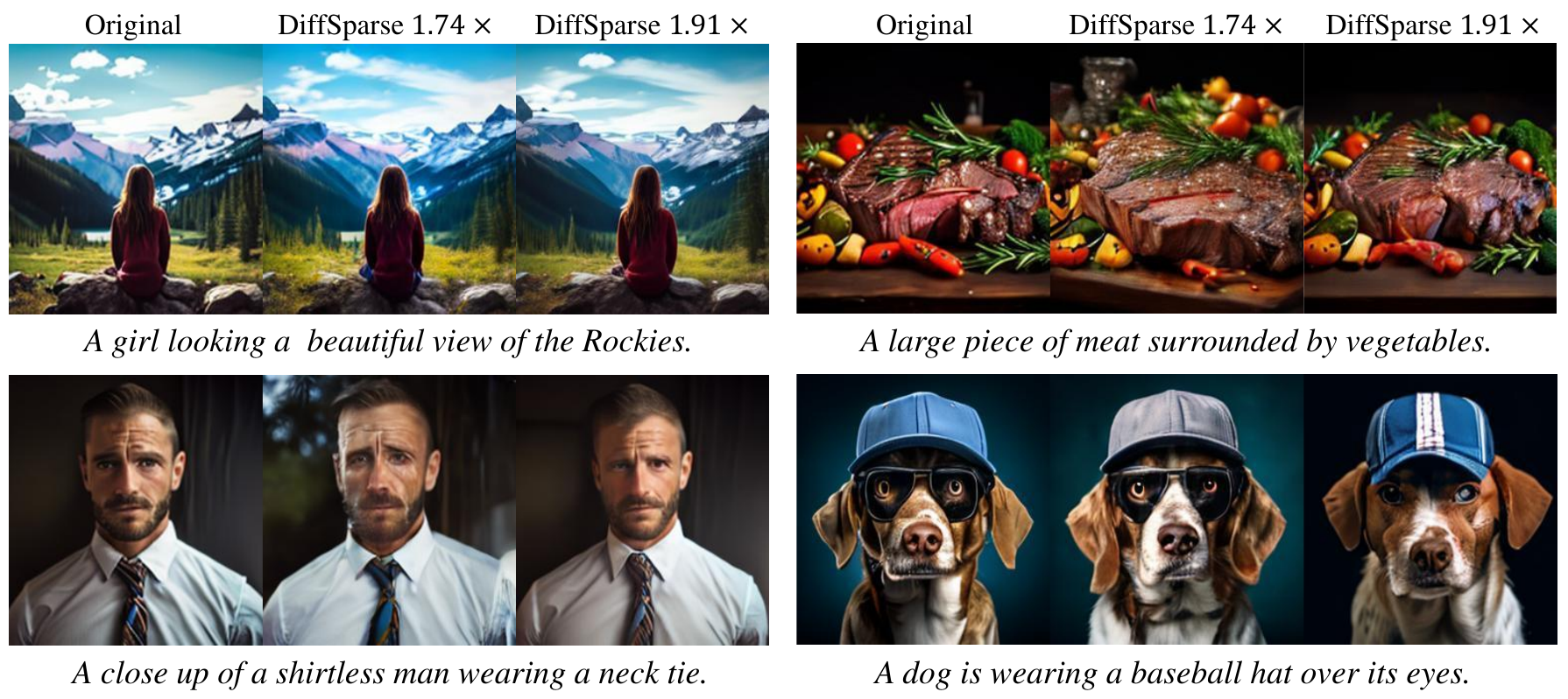}
    \caption{Comparison of our method with the baseline (PixArt-$\alpha$ with DPM-Solver++ using 20 steps) under different acceleration rates. }
    \label{fig:more_vis}
    \vspace{-10pt}
\end{figure}

\begin{figure}[h]
    \centering
    \includegraphics[width=1.0\linewidth]{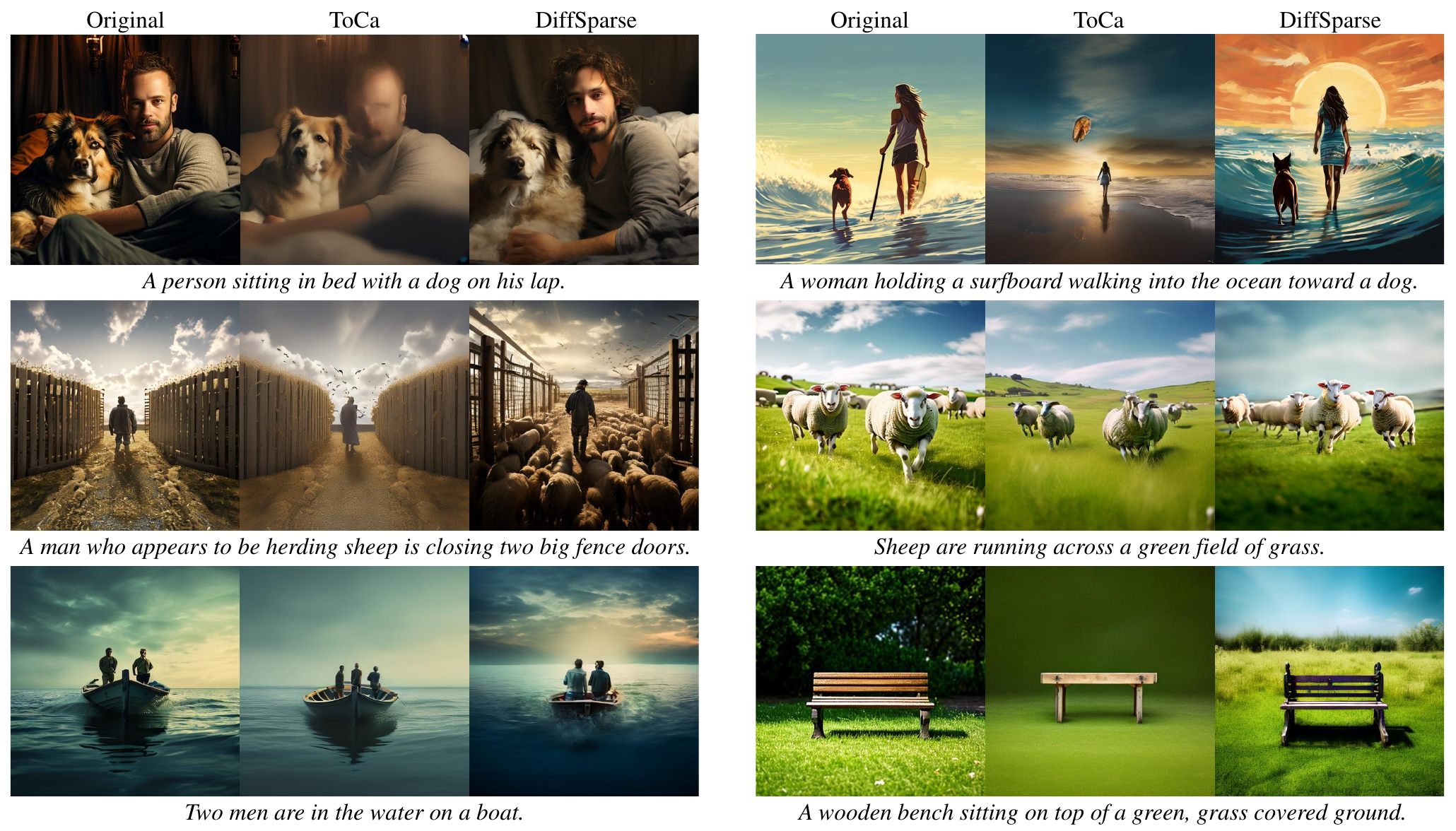}
    \caption{Comparison between our DiffSparse, and ToCa with the baseline (PixArt-$\alpha$ with DPM-Solver++ using 20 steps under 512$\times$512 resolution). }
    \label{fig:pixart_512}
\end{figure}

Figures~\ref{fig:more_vis} and~\ref{fig:pixart_512} present further visualizations, including additional comparison samples and higher-resolution results. They confirm that our approach delivers markedly higher acceleration ratios compared to the baseline, while preserving performance quality.

\end{document}